\documentclass{article} %
\usepackage{iclr2021_conference,times}
\usepackage{wrapfig,lipsum,booktabs}

\usepackage{amsmath,amsfonts,bm}

\def\eqref#1{equation~\ref{#1}}

\def\1{\bm{1}}

\DeclareMathAlphabet{\mathsfit}{\encodingdefault}{\sfdefault}{m}{sl}
\SetMathAlphabet{\mathsfit}{bold}{\encodingdefault}{\sfdefault}{bx}{n}

\usepackage{hyperref}
\usepackage{url}
\usepackage{multirow}
\usepackage{algorithm}
\usepackage[noend]{algpseudocode}
\usepackage{bm}
\usepackage{graphicx}
\usepackage{xspace} 
\usepackage{xcolor}
\usepackage{subcaption} 
\newcommand{\specified}[1]{}

\newcommand{\lrlambda}{\alpha_\lambda\xspace}
\newcommand{\rcpo}{RC-D4PG\xspace}
\newcommand{\mglr}{MetaL\xspace}
\newcommand{\mgrr}{MeSh\xspace}
\newcommand{\peq}{\mathrel{{+}{=}}}

\newcommand{\deriv}{\mathrm{d}}
\newcommand{\significant}{}

\usepackage{amsthm}
\usepackage{amssymb}

\newtheorem{theorem}{Theorem}

\title{Balancing Constraints and Rewards with\\ Meta-Gradient D4PG}

\author{Dan A. Calian\thanks{indicates equal contribution.}\\\texttt{dancalian@google.com} \And Daniel J. Mankowitz\footnotemark[1]\\\texttt{dmankowitz@google.com} \AND Tom Zahavy \And Zhongwen Xu \And Junhyuk Oh \And Nir Levine \And Timothy Mann \AND {\normalfont DeepMind}\\ London, UK}

\iclrfinalcopy %
\begin{document}
\maketitle
\begin{abstract}
Deploying Reinforcement Learning (RL) agents to solve real-world applications often requires satisfying complex system constraints. Often the constraint thresholds are incorrectly set due to the complex nature of a system or the inability to verify the thresholds offline (e.g, no simulator or reasonable offline evaluation procedure exists). This results in solutions where a task cannot be solved without violating the constraints. However, in many real-world cases, constraint violations are undesirable yet they are not catastrophic, motivating the need for soft-constrained RL approaches. We present a soft-constrained RL approach that utilizes meta-gradients to find a good trade-off between expected return and minimizing constraint violations. We demonstrate the effectiveness of this approach by showing that it consistently outperforms the baselines across four different Mujoco domains.
\end{abstract}
\section{Introduction}
Reinforcement Learning (RL) algorithms typically try to maximize an expected return objective \citep{sutton2018reinforcement}. This approach has led to numerous successes in a variety of domains which include board-games \citep{silver2017mastering}, computer games \citep{mnih2015human,tessler2017deep} and robotics \citep{abbas2018a}. However, formulating real-world problems with only an expected return objective is often sub-optimal when tackling many applied problems ranging from recommendation systems to physical control systems which may include robots, self-driving cars and even aerospace technologies. In many of these domains there are a variety of challenges preventing RL from being utilized as the algorithmic solution framework. Recently, \cite{Dulac2019} presented nine challenges that need to be solved to enable RL algorithms to be utilized in real-world products and systems. One of those challenges is handling constraints. All of the above domains may include one or more constraints related to cost, wear-and-tear, or safety, to name a few.

\textbf{Hard and Soft Constraints:} There are two types of constraints that are encountered in constrained optimization problems; namely hard-constraints and soft-constraints \citep{boyd2004convex}. Hard constraints are pairs of pre-specified functions and thresholds that require the functions, when evaluated on the solution, to respect the thresholds. As such, these constraints may limit the feasible solution set. Soft constraints are similar to hard constraints in the sense that they are defined by pairs of pre-specified functions and thresholds, however, a soft constraint does not require the solution to hold the constraint; instead, it penalizes the objective function (according to a specified rule) if the solution violates the constraint~\citep{boyd2004convex,thomas2017ensuring}.

\textbf{Motivating Soft-Constraints:} In real-world products and systems, there are many examples of soft-constraints; that is, constraints that can be violated, where the violated behaviour is undesirable but not catastrophic \citep{thomas2017ensuring,dulacarnold2020empirical}. One concrete example is that of energy minimization in physical control systems. Here, the system may wish to reduce the amount of energy used by setting a soft-constraint. Violating the constraint is inefficient, but not catastrophic to the system completing the task. In fact, there may be desirable characteristics that can only be attained if there are some constraint violations (e.g., a smoother/faster control policy).  Another common setting is where it is unclear how to set a threshold. In many instances, a product manager may desire to increase the level of performance on a particular product metric $A$, while ensuring that another metric $B$ on the same product does not drop by `approximately X$\%$'. The value `X' is often inaccurate and may not be feasible in many cases. In both of these settings, violating the threshold is undesirable, yet does not have catastrophic consequences. 

\textbf{Lagrange Optimization:} In the RL paradigm, a number of approaches have been developed to incorporate hard constraints into the overall problem formulation \citep{altman1999constrained,Tessler2018a,efroni2020explorationexploitation,achiam2017constrained,bohez2019value,chow2018lyapunovbased,Paternain2019,zhang2020reward, efroni2020explorationexploitation}. One popular approach is to model the problem as a Constrained Markov Decision Process (CMDP) \citep{altman1999constrained}. In this case, one method is to solve the following problem formulation:  ${\max_{\pi} J_{R}^{\pi}}$ s.t. $J_{C}^{\pi} \leq \beta$, where $\pi$ is a policy, $J_{R}^{\pi}$ is the expected return, $J_{C}^{\pi}$ is the expected cost and $\beta$ is a constraint violation threshold. This is often solved by performing alternating optimization on the unconstrained Lagrangian relaxation of the original problem (e.g.~\cite{Tessler2018a}), defined as: $\min_{\lambda \geq 0} \max_{\pi}  J_{R}^{\pi} + \lambda (\beta - J_{C}^{\pi})$. The updates alternate between learning the policy and the Lagrange multiplier $\lambda$.  

In many previous constrained RL works \citep{achiam2017constrained,Tessler2018a,ray2019benchmarking,satija2020constrained}, because the problem is formulated with hard constraints, there are some domains in each case where a feasible solution is not found. This could be due to approximation errors, noise, or the constraints themselves being infeasible. The real-world applications, along with empirical constrained RL research results, further motivates the need to develop a soft-constrained RL optimization approach. %
Ideally, in this setup, we would like an algorithm that satisfies the constraints while solving the task by maximizing the objective. If the constraints cannot be satisfied, then this algorithm finds a good trade-off (that is, minimizing constraint violations  while solving the task by maximizing the objective).

In this paper, we extend the constrained RL Lagrange formulation to perform soft-constrained optimization by formulating the constrained RL objective as a nested optimization problem \citep{sinha2017review} using meta-gradients. We propose \mglr that utilizes meta-gradients \citep{Xu2018,Zahavy2020} to improve upon the trade-off between reducing constraint violations and improving expected return. We focus on Distributed Distributional Deterministic Policy Gradients (D4PG) \citep{barth2018distributed} as the underlying algorithmic framework, a state-of-the-art continuous control RL algorithm. We show that \mglr can capture an improved trade-off between expected return and constraint violations compared to the baseline approaches. We also introduce a second approach called \mgrr that utilizes meta-gradients by adding additional representation power to the reward shaping function. Our \textbf{main contributions} are as follows: \textbf{(1)} We extend D4PG to handle constraints by adapting it to Reward Constrained Policy Optimization (RCPO) ~\citep{Tessler2018a} yielding Reward Constrained D4PG (RC-D4PG); \textbf{(2)} We present a soft constrained meta-gradient technique: \textbf{Meta}-Gradients for the \textbf{L}agrange multiplier learning rate (\mglr)\footnote{This is also the first time meta-gradients have been applied to an algorithm with an experience replay.}; \textbf{(3)} We derive the meta-gradient update for \mglr (Theorem 1); \textbf{(4)} We perform extensive experiments and investigative studies to showcase the properties of this algorithm. \mglr  outperforms the baseline algorithms across domains, safety coefficients and thresholds from the Real World RL suite~\citep{dulacarnold2020empirical}.

\section{Background}
A \textbf{Constrained Markov Decision Process (CMDP)} is an extension to an MDP \citep{sutton2018reinforcement} and consists of the tuple $\langle S, A, P, R, C, \gamma \rangle$ where $S$ is the state space; $A$ is the action space; $P:S\times A \rightarrow S$ is a function mapping states and actions to a distribution over next states; $R:S\times A \rightarrow \mathbb{R}$ is a bounded reward function and $C:S\times A \rightarrow \mathbb{R}^K$ is a $K$ dimensional function representing immediate penalties (or costs) relating to $K$ constraints. The solution to a CMDP is a policy $\pi:S\rightarrow \Delta_A$ which is a mapping from states to a probability distribution over actions. This policy aims to maximize the expected return $J_R^\pi=\mathbb{E}[\sum_{t=0}^\infty \gamma^{t}r_t]$ and satisfy the constraints $J_{C_{i}}^\pi=\mathbb{E}[\sum_{t=0}^\infty \gamma^{t}c_{i, t}] \leq \beta_i, i=1 \ldots K$. For the purpose of the paper, we consider a single constraint; that is, $K=1$, but this can easily be extended to multiple constraints.

\textbf{Meta-Gradients} is an approach to optimizing hyperparameters such as the discount factor, learning rates, etc.\ by performing online cross validation while simultaneously optimizing for the overall RL optimization objective such as the expected return~\citep{Xu2018,Zahavy2020}. The goal is to optimize both an inner loss and an outer loss. The update of the $\theta$ parameters on the \textit{inner loss} is defined as $\theta' = \theta + f(\tau,\theta,\eta)$, where $\theta \in \mathbb{R}^d$ corresponds to the parameters of the policy $\pi_\theta(a|s)$ and the value function  $v_{\theta}(s)$ (if applicable). The function $f:\mathbb{R}^k\rightarrow \mathbb{R}^d$ is the gradient of the policy and/or value function with respect to the parameters $\theta$ and is a function of an n-step trajectory $\tau=\langle s_1,a_1,r_2,s_2 \ldots s_n \rangle$, meta-parameters $\eta$ and is weighted by a learning rate $\alpha$ and is defined as $f(\tau,\theta,\eta) = \alpha \frac{\deriv J_{obj}^{\pi_{\theta}}(\theta, \tau, \eta)}{\deriv\theta}$ where $J_{obj}^{\pi_{\theta}}(\theta, \tau, \eta)$ is the objective being optimized with respect to $\theta$. The idea is to then evaluate the performance of this new parameter value $\theta'$ on an \textit{outer loss} -- the meta-gradient objective. We define this objective as $J'(\tau',\theta',\bar{\eta})$ where $\tau'$ is a new trajectory, $\theta'$ are the updated parameters and $\bar{\eta}$ is a \textit{fixed} meta-parameter (which needs to be selected/tuned in practice). We then need to take the gradient of the objective $J'$ with respect to the meta-parameters $\eta$ to yield the outer loss update $\eta' = \eta + \alpha_\eta \frac{\partial J'(\tau',\theta',\bar{\eta})}{\partial \eta}$. This gradient is computed as follows: $\frac{\partial J'(\tau',\theta',\bar{\eta})}{\partial \eta} = \frac{\partial J'(\tau',\theta',\bar{\eta})}{\partial \theta'} \frac{\partial \theta'}{\partial \eta}$. The outer loss is essentially the objective we are trying to optimize. This could be a policy gradient loss, a temporal difference loss, a combination of the two etc~\citep{Xu2018,Zahavy2020}. Meta-gradients have been previously used to learn intrinsic rewards for policy gradient~\citep{zheng2018learning} and auxiliary tasks~\citep{veeriah2019discovery}. Meta-gradients have also been used to adapt optimizer parameters~\citep{young2018metatrace,franceschi2017forward}. In our setup, we consider the continuous control setting, provide the first implementation of meta-gradients for an algorithm that uses an experience replay, and focus on adapting meta-parameters that encourage soft constraint satisfaction while maximizing expected return.

\textbf{D4PG} is a state-of-the-art continuous control RL algorithm with a deterministic policy \citep{barth2018distributed}. It is an incremental improvement to DDPG~\citep{lillicrap2015continuous}. The overall objective of DDPG is to maximize $J(\theta_{a},\theta_{c}) = \mathbb{E}[ Q_{\theta_{c}}(s,a) | s=s_t, a = \pi_{\theta_{a}}(s_t)]$ where $\pi_{\theta_{a}}(s_t)$ is a deterministic policy with parameters $\theta_{a}$ and $Q_{\theta_{c}}(s,a)$ is an action value function with parameters $\theta_{c}$. The actor loss is defined as: 
${L_{actor} = \Vert \textrm{SG}(\nabla_a Q_{\theta_{c}}(s_t,a_t)|_{a_t=\pi_{\theta_{a}}(s)} + a_{\theta_{a},t}) - a_{\theta_{a},t} \Vert_2}$ where SG is a stop gradient. The corresponding gradient update is defined as ${\nabla_{\theta_{a}} J(\theta_{a}) = \mathbb{E}[ \nabla_a Q_{\theta_{c}}(s,a) \nabla_{\theta_{a}} \pi_{\theta_{a}}(s_t)]}$. The critic is updated using the standard temporal difference error loss: 
${L_{critic} = (r(s,a) + \gamma Q_{T}(s',\pi_{T}(s')) - Q_{\theta_{c}}(s,a))^2}$ where $Q_{T}, \pi_{T}$ are the target critic and actor networks respectively. In D4PG, the critic is a distributional critic based on the C51 algorithm \citep{bellemare2017distributional} and the agent is run in a distributed setup with multiple actors executed in parallel, n-step returns and with prioritized experience replay. We will use the non-distributional critic update in our notation for ease of visualization and clarity for the reader\footnote{This can easily be extended to include the distributional critic.}.
\section{Reward Constrained D4PG (RC-D4PG)}
This section describes our modifications required to transform D4PG into Reward Constrained D4PG (\rcpo) such that it maximizes the expected return and satisfies constraints.

The constrained optimisation objective is defined as: $\max_{\pi_{\theta}} J_{R}^{\pi_{\theta}}$ subject to $J_{C}^{\pi_{\theta}} \leq \beta$, where ${J_{R}^{\pi_{\theta}} = \mathbb{E}[ Q(s,a) | s=s_t, a = \pi_{\theta}(s_t) ]}$ and ${J_{C}^{\pi_{\theta}} = \mathbb{E}[ C(s,a) | s=s_t, a = \pi_{\theta}(s_t) ]}$; the parameter ${\theta=\langle \theta_a, \theta_c \rangle}$ from here on in; $C(s,a)$ is a long-term penalty value function (e.g., sum of discounted immediate penalties) corresponding to constraint violations. The Lagrangian relaxation objective is defined as ${J_{R}^{\pi_{\theta}} + \lambda ( \beta - J_{C}^{\pi_{\theta}})}$. As in RCPO, a proxy objective ${J_{R}^{\pi_{\theta}} - \lambda J_{C}^{\pi_{\theta}}}$ is used that converges to the same set of locally optimal solutions as the relaxed objective \citep{Tessler2018a}. Note that the constant $\beta$ does not affect the policy improvement step and is only used for the Lagrange multiplier loss update. To optimize the proxy objective with D4PG, reward shaping of the form $r(s,a) - \lambda c(s,a)$ is required to yield the reward shaped critic loss defined as: ${L_{critic}(\theta_{c}, \lambda) = (r(s,a) - \lambda c(s,a) + \gamma Q_{T}(s,\pi_{T}(s')) - Q_{\theta_{c}}(s,a)})^2$. The actor loss is defined as before. The Lagrange loss is defined as: ${L_{lagrange}(\lambda)= \lambda (\beta - J_C^{\pi_{\theta}})}$ where $\lambda \geq 0$. Since RC-D4PG is off-policy, it requires storing the per time-step penalties, $c$, inside the transitions stored in the experience replay buffer (ER). For training the Lagrange multiplier an additional penalty buffer is used to store the per-episode penalties $J_C^{\pi_{\theta}}$. The learner then reads from this penalty buffer for updating the Lagrange multiplier. RC-D4PG updates the actor/critic parameters and the Lagrange multipliers using alternating optimization. The full algorithm for this setup can be found in the Appendix, Algorithm \ref{alg:rcd4pg}.

\section{Meta-Gradients for the Lagrange  learning rate (\mglr)}
In this section, we introduce the \mglr algorithm which extends \rcpo to use meta-gradients for adapting the learning rate of the Lagrangian multiplier. The idea is to update the learning rate such that the outer loss (as defined in the next subsection) is minimized.
Our intuition is that a learning rate gradient that takes into account the overall task objective and constraint thresholds will lead to improved overall performance.

\begin{algorithm}[H]
\caption{\mglr}\label{alg:mlrc}
\begin{algorithmic}[1]
\State \textbf{Input:}  penalty $c(\cdot)$, constraint $C(\cdot)$, threshold $\beta$, learning rates $\alpha_1, \alpha_{\theta_{a}}, \alpha_{\theta_{c}}, \alpha_{\eta}$, max.\ number of episodes $M$ 
\State Initialize actor, critic parameters  $\theta_{a}$ and $\theta_c$, Lagrange multiplier $\lambda = 0$, meta-parameter $\eta={\alpha_\lambda}$
    \For{$1 \ldots M$}
        \State \textbf{Inner loss: }
        \State Sample episode penalty $J_C^{\pi}$ from the penalty replay buffer
        \State $\lambda' \gets  [\lambda - \alpha_1 \exp(\alpha_\lambda) \left(\beta - J_C^{\pi} \right)]_{+} $ \Comment{\textbf{Lagrange multiplier update}}
        \State Sample a batch with tuples $\langle s_t, a_t, r_t, c_t \rangle_{t=1}^{T}$ from ER and split into training/validation sets
        \State Accumulate and apply actor and critic updates over training batch $T_{train}$ by:
        \State $\nabla \theta_c = 0, \nabla \theta_a = 0$
            \For{$t = 1 \ldots T_{train}$}
            \State $\hat R_t = r_t - \lambda' c_t + \gamma \hat Q(\lambda, s_{t+1}, a_{t+1}\sim \pi_T(s_{t+1}); \theta_{c})$
            \State $\nabla \theta_c \peq \alpha_{\theta_{c}} \partial (\hat{R}_t - \hat Q(\lambda, s_t, a_t; \theta_{c}))^2 / \partial \theta_{c}$ \Comment{\textbf{Critic update}}
            \State $\nabla \theta_a \peq \alpha_{\theta_{a}}  \mathbb{E} [\nabla_a Q(s_t,a_t) \nabla_{\theta_{a}} \pi_{\theta_{a}}(s_t)|_{a_t=\pi(s_t)}]$ \Comment{\textbf{Actor update}}
            \EndFor
        \State $\theta'_{c} \gets \theta_{c} - \frac{1}{T_{train}}\sum \nabla \theta_c$
        \State $\theta'_{a} \gets \theta_{a} + \frac{1}{T_{train}}\sum \nabla \theta_a$
        \State \textbf{Outer loss: } Compute outer loss and meta-gradient update using validation set $T_{validate}$:
        \State $\lrlambda' \gets \lrlambda - \alpha_\eta \frac{\partial J'(\theta'_{c}(\lrlambda), \lambda'(\lrlambda))}{\partial \lrlambda}$  \Comment{\textbf{Meta-parameter update - Theorem 1}}
        \State $\lambda\gets\lambda',\theta_{a}\gets\theta'_{a}, \theta_{c}\gets\theta'_{c}, \lrlambda\gets\lrlambda'$
    \EndFor
\State \textbf{return} $\theta_{a}, \theta_{c}, \lambda$ 
\end{algorithmic}
\label{alg:mlrcd4pg}
\end{algorithm}
\paragraph{Meta-parameters, inner and outer losses:} The meta-parameter is defined as $\eta=\lrlambda$. The \textit{inner loss} is composed of three losses, the actor, critic and Lagrange loss respectively. The actor and critic losses are the same as in RC-D4PG. The Lagrange multiplier loss is defined as:
${L_{lagrange}(\lambda)=\exp(\lrlambda)\lambda (\beta - J_C^{\pi})}$ where $\lrlambda$ is the meta-parameter as defined above. The meta-parameter is wrapped inside an exponential function to magnify the effect of $\lrlambda$ while also ensuring non-negativity of the effective learning rate. The inner loss updates are
${
\begin{bmatrix}
\theta_{a}'\\
\theta_{c}' \\
\lambda'
\end{bmatrix}
= 
\begin{bmatrix}
\theta_{a}\\
\theta_{c} \\
\lambda 
\end{bmatrix}
-
\begin{bmatrix}
f(\tau, \theta_a, \eta)\\
f(\tau, \theta_c, \eta)\\
f(\tau, \lambda, \eta) 
\end{bmatrix}
=
\begin{bmatrix}
\theta_{a}\\
\theta_{c} \\
\lambda 
\end{bmatrix}
-
\begin{bmatrix}
\alpha_{\theta_{a}} \frac{d L_{actor}(\theta_a)}{d \theta_a}\\
\alpha_{\theta_{c}} \frac{d L_{critic}(\theta_c, \lambda)}{d \theta_c}\\
\alpha_1 \frac{d L_{lagrange}(\lambda)}{d \lambda} 
\end{bmatrix}}
$
where $\alpha_{\theta_{a}}, \alpha_{\theta_{c}}, \alpha_1$ are the fixed actor critic and Lagrange multiplier learning rates respectively. The \textit{outer loss} is defined as $J'(\theta'_{c}(\lrlambda), \lambda'(\lrlambda)) = L_{outer} = L_{critic}(\theta'_{c}(\lrlambda),\lambda'(\lrlambda))$. We tried different variants of outer losses and found that this loss empirically yielded the best performance; we discuss this in more detail in the experiments section.
This is analogous to formulating \mglr as the following nested optimization problem: 
${\min_{\alpha_{\lambda}} J'(\theta(\alpha_{\lambda}), \lambda(\alpha_{\lambda})), \mbox{ s.t. } \theta, \lambda  \in \arg\min_{\theta, \lambda \geq 0} \{ {-J_{R}^{\pi_{\theta}} - \lambda (\lrlambda) (\beta - J_{C}^{\pi_{\theta}}})\}}
$. We treat the lower level optimization problem as the Lagrange relaxation objective (inner loss). We then treat the upper level optimization as the meta-gradient objective $J'(\theta(\lrlambda), \lambda(\alpha_{\lambda}))$ (outer loss). This transforms the optimization problem into soft-constrained optimization since the meta-parameter $\lrlambda$ guides the learning of the Lagrange multiplier $\lambda$ to minimize the outer loss while attempting to find a good trade-off between minimizing constraint violations and maximizing return (inner loss). 

As shown in Algorithm \ref{alg:mlrcd4pg}, the inner loss gradients are computed for $\lambda$ (line 6), $\theta_c$ (line 12) and $\theta_a$ (line 13) corresponding to the Lagrange multiplier, critic and actor parameters respectively. The Lagrange multiplier is updated by sampling episode penalties which is an empirical estimate of $J_C^{\pi}$ from a separate penalty replay buffer (line 5) to compute the gradient update. The updated multiplier is then utilized in the critic inner update (lines 11 and 12) to ensure that the critic parameters are a function of this new updated Lagrange multiplier. The actor and critic parameters are updated using the training batch, and these updated parameters along with a validation batch are used to compute the outer loss (line 17). The meta-parameter $\lrlambda$ is then updated along the gradient of this outer loss with respect to $\eta=\lrlambda$. We next derive the meta-gradient update for $\alpha_\lambda$, and present it in the following theorem (see the Appendix, Section \ref{sec:metal_derivation} for the full derivation). Intuition for this meta-gradient update is provided in the experiments section.
\begin{theorem}{\mglr gradient update:}
Let $\beta \geq 0$ be a pre-defined constraint violation threshold, meta-parameter $\eta=\lrlambda$ and $J_C^{\pi_{\theta}}=\mathbb{E}\biggl[ C(s,a) | s=s_t, a =\pi_{\theta}(s_t) \biggr]$ is the discounted constraint violation function, then,
the meta-gradient update is: 
\begin{equation*}
  \lrlambda' \leftarrow \lrlambda - \alpha_\eta \biggl( -2\delta \cdot c(s,a) \cdot \alpha_1 \exp(\lrlambda) \cdot\biggl(J_C^{\pi_{\theta}} - \beta \biggr) \biggl( -2 \alpha_{\theta_{c}}  (\nabla_{\theta'_{c}} Q_{\theta'_{c}}(s,a))^T  \nabla_{\theta_{c}} Q_{\theta_c}(s,a) + 1 \biggr)\biggr) \enspace ,
\end{equation*}

where $\delta$ is the TD error; $\alpha_{\theta_{c}}$ is the critic learning rate and $\alpha_\eta$ is the meta-parameter learning rate. 
\end{theorem}

\section{Experiments}
The experiments were performed using domains from the Real-World Reinforcement Learning (RWRL) suite\footnote{\url{https://github.com/google-research/realworldrl_suite}}, namely cartpole:swingup, walker:walk, quadruped:walk and humanoid:walk. We will refer to these domains as cartpole, walker, quadruped and humanoid from here on in.

We focus on two types of tasks with constraints: (1) solvable constraint tasks - where the task is solved and the constraints can be satisfied; (2) unsolvable constraint tasks - where the task can be solved but the constraints \textit{cannot} be satisfied. Unsolvable constraint tasks correspond to tasks where the constraint thresholds are incorrectly set and cannot be satisfied, situations which occur in many real-world problems as motivated in the introduction. The specific constraints we focused on for each domain can be found in the Appendix (Section \ref{app:constraint_defs}). The goal is to showcase the soft-constrained performance of \mglr, with respect to reducing constraint violations and maximizing the return in both of these scenarios (solvable and unsolvable constraint tasks) with respect to the baselines.

The baseline algorithms we focused on for each experiment are D4PG without any constraints, RC-D4PG (i.e., hard constraint satisfaction) and Reward Shaping D4PG (RS-D4PG) (i.e., soft constraint satisfaction). RS-D4PG uses a fixed $\lambda$ for the duration of training. We compare these baselines to \mglr. Note that D4PG,  RC-D4PG and \mglr have \textit{no prior information} regarding the Lagrange multiplier. \rcpo and \mglr attempt to learn a suitable multiplier value from scratch, i.e.\ the initial Lagrange multiplier value is set to $0.0$. In contrast, RS-D4PG has prior information (i.e.\ it uses a pre-selected fixed Lagrange multiplier). 

\textbf{Experimental Setup}: For each domain, the action and observation dimensions are shown in the Appendix, Table~\ref{tab:state_dimensions}. The episode length is $1000$ steps, the base reward function is computed within the \texttt{dm\_control} suite~\citep{Tassa2018}. The upper bound reward for each task is $1000$. Each task was trained for $20000$ episodes. Each variant of D4PG uses the same network architecture (see the Appendix, Table~\ref{tab:d4pg_hyperparameters} for more details).  %

We use different performance metrics to compare overall performance. We track the average episode return ($R$), but we also define the \textbf{penalized return}: $R_{penalized}=R - \kappa \cdot \psi_{\beta,C}$, which captures the trade-off between achieving optimal performance and satisfying the constraints. Here, $R$ is the average return for the algorithm upon convergence (computed as an average over the previous $100$ episodes); $\kappa$ is a fixed constant that determines how much to weight the constraint violation penalty. For the purposes of evaluation, we want to penalize algorithms that consistently violate the constraints and therefore set $\kappa=1000$. Since the upper bound of rewards for each domain is $1000$, we are essentially weighing equally attaining high performance and satisfying constraints. Finally, $\psi_{\beta,C}=\max(0, J_C^{\pi} - \beta)$ is defined as the \textbf{overshoot}. Here $\beta$ is the \textbf{constraint violation threshold} and defines the allowable average constraint violations per episode; $J_C^{\pi}$ is the average constraint violation value per episode upon convergence for a policy $\pi$. The overshoot, $\psi_{\beta,C}$, tracks the average constraint violations that are above the allowed constraint violation threshold $\beta$.

We investigate each algorithm's performance along a variety of dimensions which include different constraint violation thresholds (see the Appendix, Table \ref{tab:safe_envs_full} for a list of thresholds used), safety coefficients and domains. The \textit{safety coefficient} is a flag in the RWRL suite~\citep{dulac2020empirical}. This flag contains values between $0.0$ and $1.0$. Reducing the value of the flag ensures that more constraint violations occur per domain per episode. As such, we searched over the values $\{0.05, 0.1, 0.2, 0.3\}$. These values vary from solvable constraint tasks (e.g., $0.3$) to unsolvable constraint tasks (e.g., $0.05$). We wanted to see how the algorithms behaved in these extreme scenarios. In addition, we analysed the performance across a variety of different constraint violation thresholds (see Appendix, Table~\ref{tab:thresholds}). All experiments are averaged across $8$ seeds.

\subsection{Main Results}

We begin by analyzing the performance of our best  variant, \mglr, with different outer losses. Then we analyse the overall performance of all methods, followed by dissecting performance along the dimensions of safety coefficient and domain respectively. Finally, we investigate the derived gradient update for \mglr from Theorem 1 and provide intuition for the algorithm's behaviour.

\begin{figure}[!htb]%
\centering
        \centering
        \includegraphics[width=0.5\linewidth]{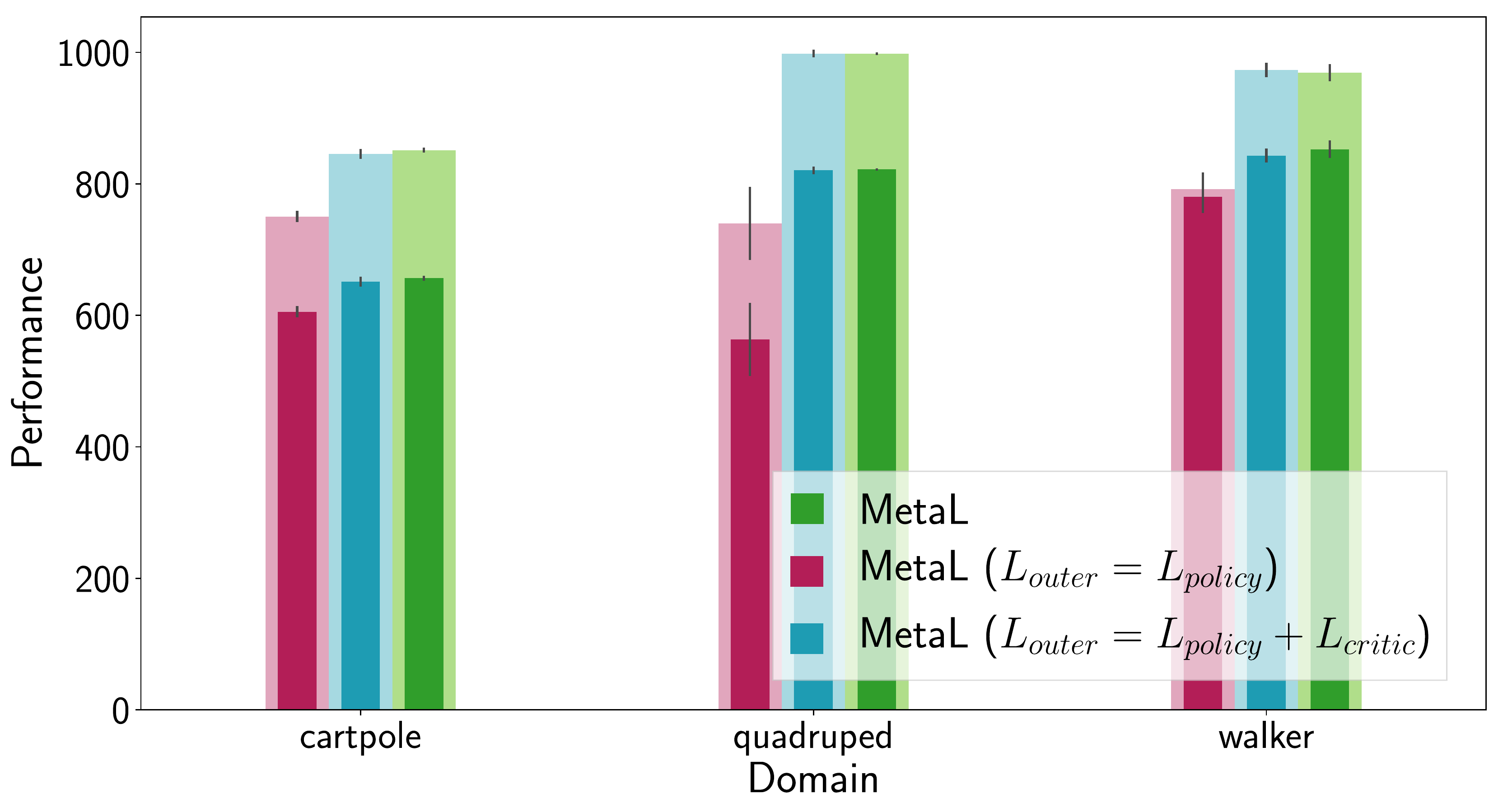}%
        \includegraphics[width=0.5\linewidth]{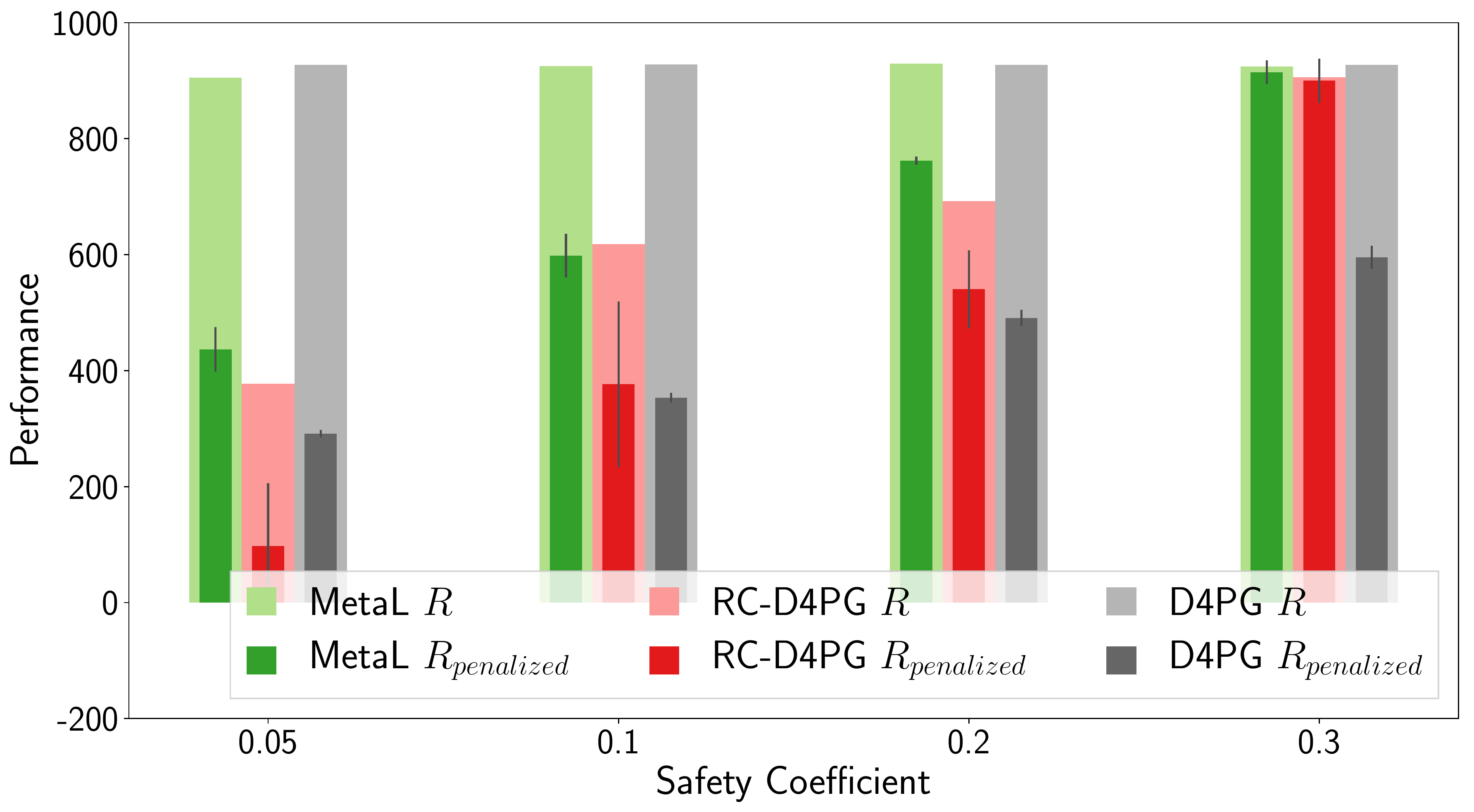}
  \caption{\mglr performance using different outer losses (left) and comparison with D4PG and RC-D4PG (right).}
  \label{fig:metalmesh}
\end{figure}

\textbf{\mglr outer loss}: We wanted to determine whether different outer losses would result in improved overall performance. We used the actor loss ($L_{actor}$) and the combination of the actor and critic losses as the outer loss ($L_{actor} + L_{critic}$) and compared them with the original \mglr outer loss ($L_{critic}$) as well as the other baselines. Figure~\ref{fig:metalmesh} shows that using just the actor loss results in the worst performance; while using the critic loss always results in better performance. The best performance is achieved by the original critic-only \mglr outer loss. 

There is some intuition for choosing a critic-only outer loss. In \mglr, the critic loss is a function of lambda. As a result, the value of lambda affects the agents ability to minimize this loss and therefore learn an accurate value function. In D4PG, an accurate value function (i.e., the critic) is crucial for learning a good policy (i.e., the actor). This is because the policy relies on an accurate estimate of the value function to learn good actions that maximize the return (see D4PG actor loss). This would explain why adding the actor loss to the outer loss does not have much effect on the final quality of the solution. However, removing the critic loss has a significant effect on the overall solution.

\textbf{Overall performance}: We averaged the performance of \mglr across all safety coefficients, thresholds and domains and compared this with the relevant baselines. As seen in Table \ref{tab:main_results}, \mglr outperforms all of the baseline approaches by achieving the best trade-off of minimizing constraint violations and maximizing return\footnote{MetaL's penalized reward ($R_{penalized}$) performance is significantly better than the baselines with all p-values smaller than $10^{-9}$ using Welch's t-test.}. This includes all of the soft constrained optimization baselines (i.e., RS-D4PG variants), D4PG as well as the hard-constrained optimization algorithm RC-D4PG. It is interesting to analyze this table to see that the best reward shaping variants are (1)  $RS-0.1$ which achieves comparable return, but higher overshoot and therefore lower penalized return; (2) $RS-1.0$ which attains significantly lower return but lower overshoot resulting in lower penalized return. D4PG has the highest return, but this results in significantly higher overshoot. While RC-D4PG attains lower overshoot, it also yields significantly lower overall return. We now investigate this performance in more detail by looking at the performance per safety coefficient and per domain.

\begin{table}
\vspace{-0.3cm}
\centering
\begin{tabular}{lccc}
\toprule
Algorithm &     $R_{penalized}$ &    $R$ &  {\scriptsize$\max(0, J_C^\pi - \beta)$} \\
\midrule%
           D4PG &           432.70 $\pm$ 11.99 &  \textbf{927.66} &           0.49 \\
 \textbf{MetaL} &  \textbf{677.93 $\pm$ 25.78} &           921.16 &           0.24 \\
        RC-D4PG &           478.60 $\pm$ 89.26 &           648.42 &  \textbf{0.17} \\
         RS-0.1 &           641.41 $\pm$ 26.67 &           906.76 &           0.27 \\
         RS-1.0 &           511.70 $\pm$ 15.50 &           684.30 &           0.17 \\
        RS-10.0 &           208.57 $\pm$ 61.46 &           385.42 &           0.18 \\
       RS-100.0 &           118.50 $\pm$ 62.54 &           314.93 &           0.20 \\
\bottomrule
\end{tabular}%
\caption{\label{tab:main_results}Overall performance across domains, safety coefficients and thresholds.}%
\vspace{-0.5cm}
\end{table}

\begin{figure}[!htb]
    \centering
        \includegraphics[width=1.0\linewidth]{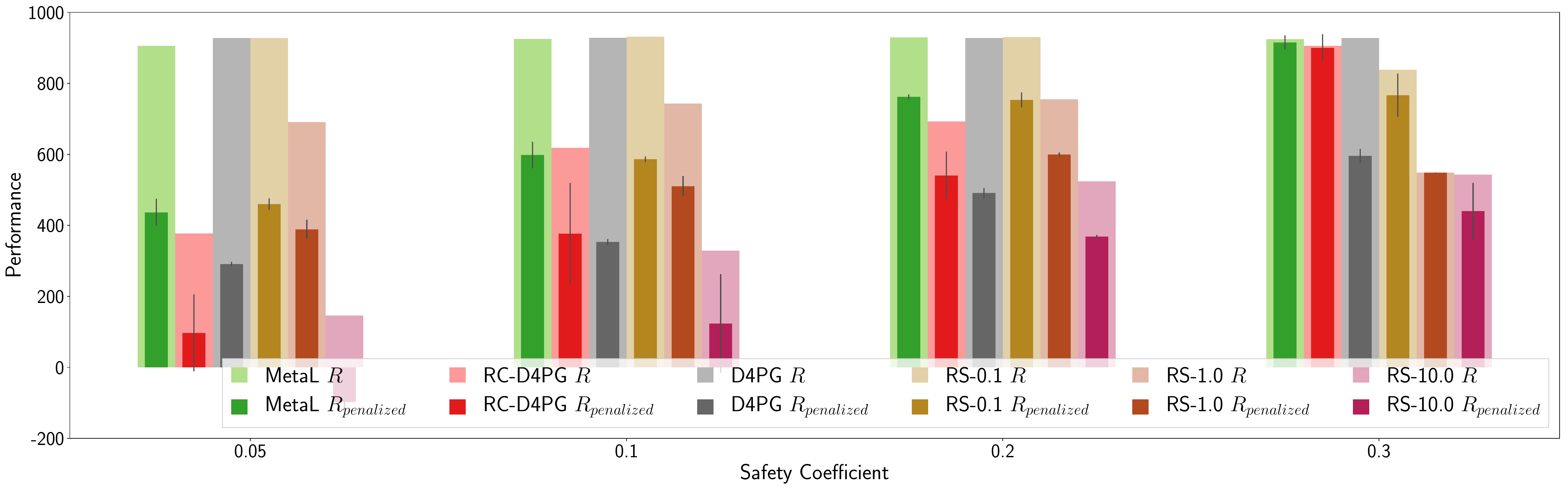}
    \caption{Performance as a function of safety coefficient.}
    \label{fig:safety_coeff}
    \vspace{-0.3cm}
\end{figure}

\begin{figure}[!htb]
    \centering
        \centering
        \includegraphics[width=1.0\linewidth]{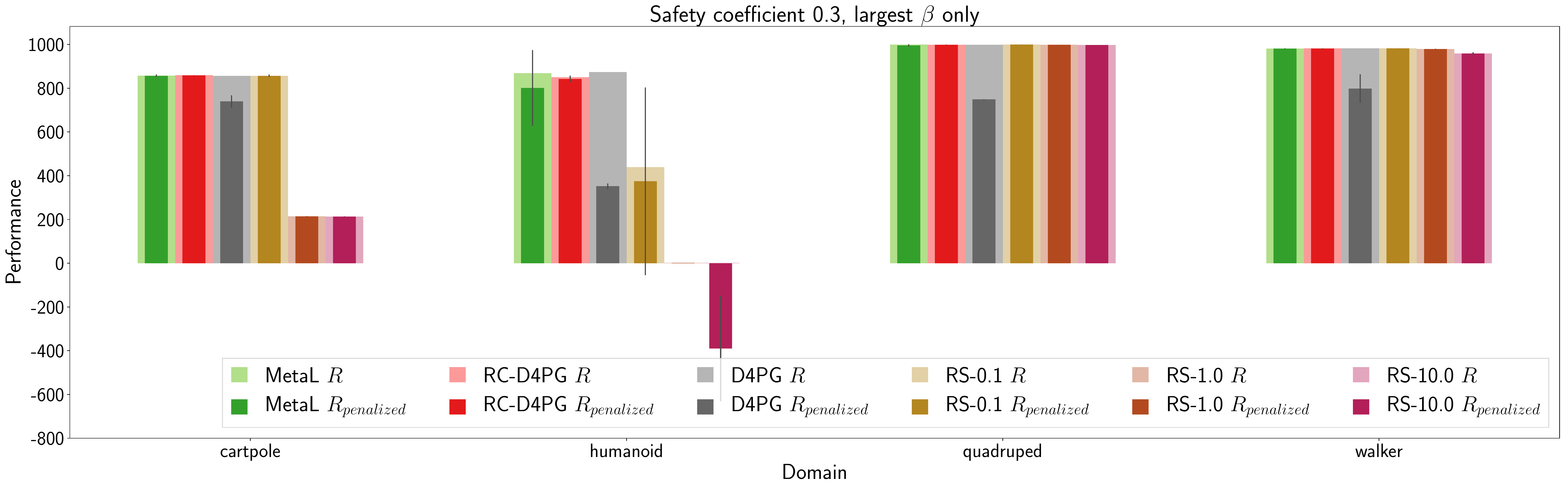}
    \caption{Performance per domain. \mglr compared to baselines in terms of average reward and penalized reward across the highest safety coefficient and largest thresholds for each domain.}
    \label{fig:domains}
    \vspace{-0.3cm}
\end{figure}

\textbf{Performance as a function of safety coefficient:} We analyzed the average performance per safety coefficient, while averaging across all domains and thresholds. As seen in Figure~\ref{fig:metalmesh} (right), \mglr achieves comparable average return to that of D4PG. In addition, it significantly outperforms both D4PG and \rcpo in terms of penalized return. Figure \ref{fig:safety_coeff} includes the reward shaping baselines. As can be seen in this figure, choosing a different reward shaping value can lead to drastically different performance. This is one of the drawbacks of the RS-D4PG variants. It is possible however, to find comparable RS variants (e.g., $RS-0.1$ for the lowest safety coefficient of $0.05$). However, as can be seen in Figure \ref{fig:domains}, for the highest safety coefficient and largest threshold, this RS variant fails completely at the humanoid task, further highlighting the instability of the RS approach. Figure \ref{fig:domains} which presents the performance of \mglr and the baselines on the highest safety coefficient and largest threshold (to ensure that the constraint task is solvable), shows that \mglr has comparable performance to \rcpo (a hard constrained optimization algorithm). This further highlights the power of \mglr whereby it can achieve comparable performance when the constraint task is solvable compared to hard constrained optimization algorithms and state-of-the-art performance when the constraint task is not solvable. 

\textbf{Performance per domain:} When analyzing the performance per domain, averaging across safety coefficients and constraint thresholds, we found that \mglr has significantly better penalized return compared to D4PG and \rcpo across the domains. A table of the results can be seen in the Appendix, Figure \ref{tab:domains}. Note that, as mentioned previously, the RS-D4PG variants fluctuate drastically in performance across domains. %

\textbf{Algorithm behaviour analysis}: Since \mglr is a soft-constrained adaptation of \rcpo, we next analyze \mglr's gradient update in Theorem 1 to understand why the performance of \mglr differs from that of \rcpo in two types of scenarios: (1) solvable and (2) unsolvable constraint tasks. For both scenarios, we investigate the performance on cartpole for a constraint threshold of $0.115$\footnote{This threshold was chosen as varying the safety coefficient at this threshold yields both solvable and unsolvable constraint tasks which is important for our analysis.}. 

\begin{figure}[!htb]
    \centering
    \begin{minipage}{.32\textwidth}
        \centering
        \includegraphics[width=\linewidth]{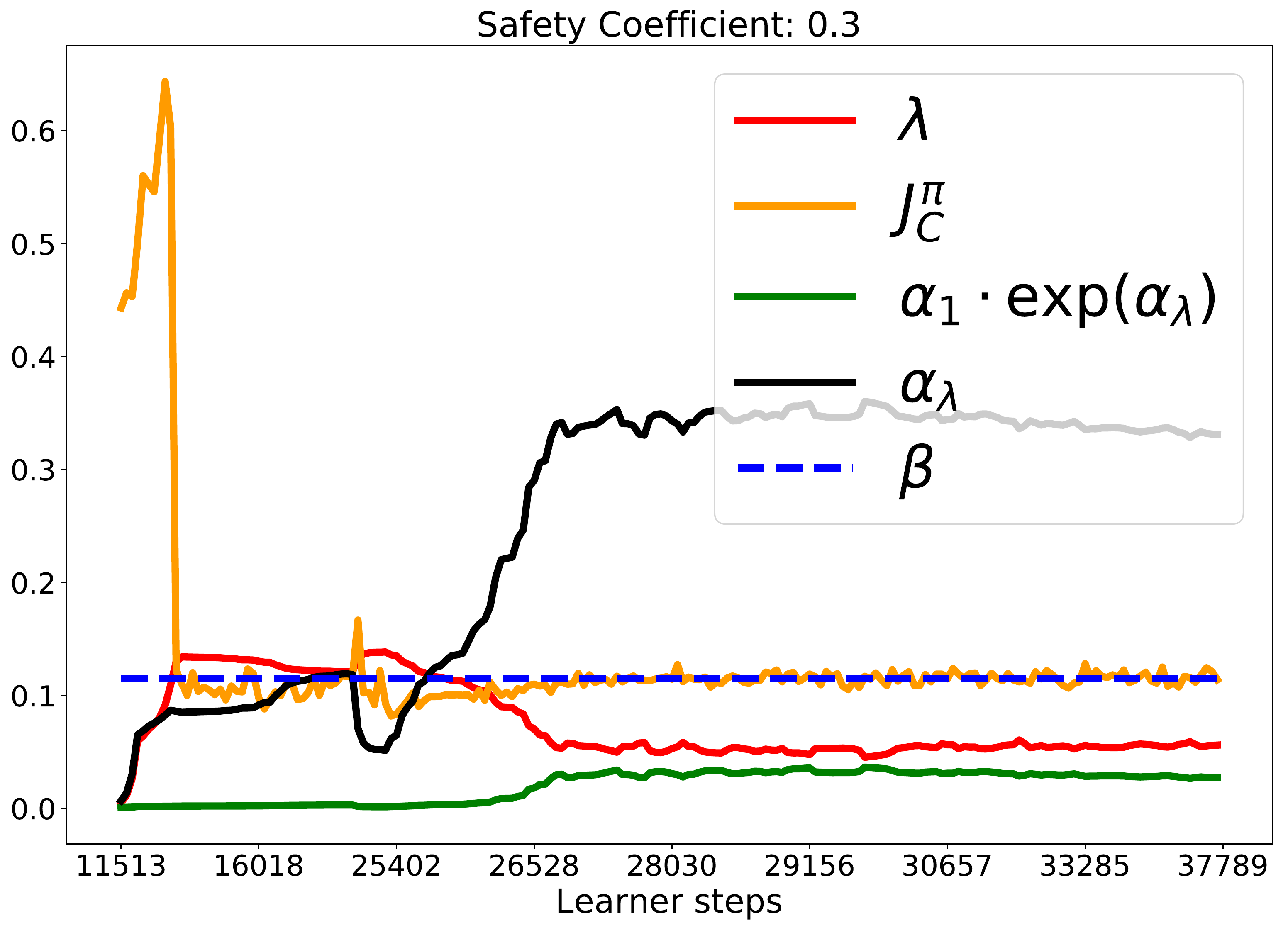}
    \end{minipage}
    \begin{minipage}{0.32\textwidth}
        \centering
        \includegraphics[width=\linewidth]{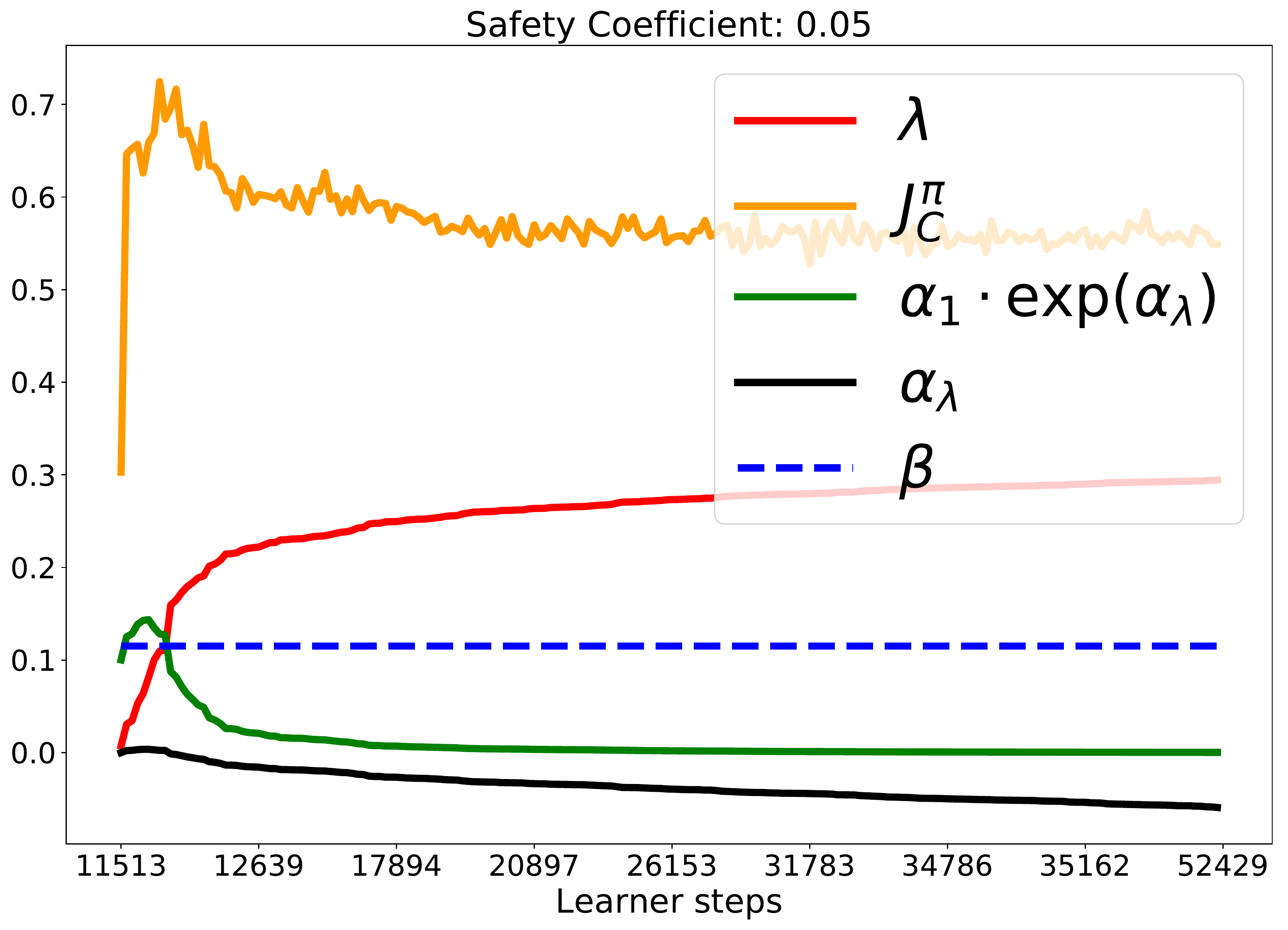}
    \end{minipage}
    \begin{minipage}{.32\textwidth}
        \centering
        \includegraphics[width=\linewidth]{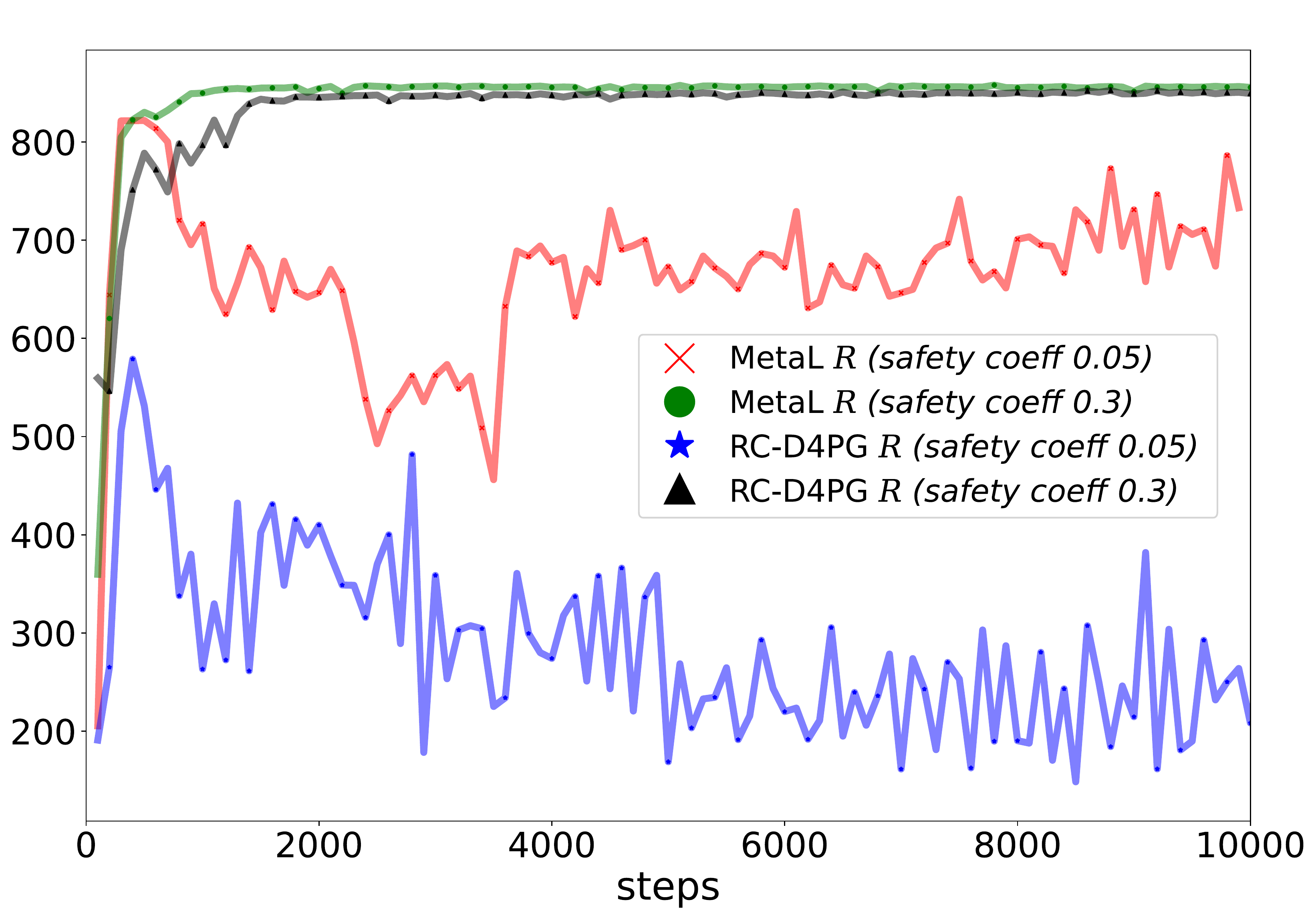}
    \end{minipage}
    \caption{The learning progress of \mglr for solvable (left) and unsolvable (middle) constraint tasks. In both cases, \mglr attempts to try and maximize the return (right).}
    \label{fig:analysis} 
    \vspace{-0.6cm}
\end{figure}

For (1), we set the safety coefficient to a value of $0.3$. The learning curve for this converged setting can be seen in Figure \ref{fig:analysis} (left). We track $4$ different parameters here: the Lagrangian multiplier $\lambda$ (red curve), the mean  penalty value $J_C^{\pi}$ (orange curve), the meta-parameter $\lrlambda$ (black curve) and the scaled Lagrangian learning rate $\alpha_1 \cdot \exp(\alpha_\lambda)$ (green curve). The threshold $\beta$ is shown as the blue dotted line. Initially there are many constraint violations. This corresponds to a large difference for $J_C^{\pi} - \beta$ (orange curve minus blue dotted line) which appears in the gradient in Theorem 1. As a result, the meta-parameter $\lrlambda$ increases in value as seen in the figure, and therefore increases the scaled learning rate to modify the value of $\lambda$ such that an improved solution can be found. Once $J_C^{\pi}$ is satisfying the constraint in expectation ($J_C^{\pi} - \beta \approx 0$), the scaled learning rate drops in value due to $J_C^{\pi} - \beta$ being small. This is an attempt by the algorithm to slow down the change in $\lambda$ since a reasonable solution has been found (see the return for \mglr (green curve) in Figure \ref{fig:analysis} (right)).

For (2), we set the safety coefficient to a  value of $0.05$ making the constraint task unsolvable in this domain. The learning curves can be seen in Figure \ref{fig:analysis} (middle). Even though the constraint task is unsolvable, \mglr still manages to yield a reasonable expected return as seen in Figure \ref{fig:analysis} (right). This is compared to \rcpo that overfits to satisfying the constraint and, in doing so, results in poor average reward performance. This can be seen in Figure \ref{fig:analysis} (middle) where \rcpo has lower overshoot than MetaL for low safety coefficients. However, this is at the expense of poor expected return and penalized return performance as seen in Figure \ref{fig:analysis} (left). We will now provide some intuition for \mglr performance and relate it to the $\alpha_\lambda$ gradient update.

In this setting, there are consistent constraint violations leading to a large value for $J_C^{\pi}-\beta$. At this point an interesting effect occurs. The value of $\lrlambda$ decreases, as seen in the figure, while it tries to adapt the value of $\lambda$ to satisfy the constraint. However, as seen in the gradient update, there is an exponential term $\exp(\lrlambda)$ which scales the Lagrange multiplier learning rate. This quickly drives the gradient down to $0$, and consequently the scaled Lagrange multiplier learning rate too, as seen in Figure~\ref{fig:analysis} (middle). This causes $\lambda$ to settle on a value as seen in the figure. At this point the algorithm optimizes for a stable fixed $\lambda$ and as a result finds the best trade-off for expected return at this $\lambda$ value. In summary, \mglr will maximize the expected return for an `almost' fixed $\lambda$, whereas \rcpo will attempt to overfit to satisfying the constraint resulting in a poor overall solution.

\section{Discussion}
In this paper, we presented a soft-constrained RL technique called \mglr that combines meta-gradients and constrained RL to find a good trade-off between minimizing constraint violations and maximizing returns. This approach (1) matches the return and constraint performance of a hard-constrained optimization algorithm (RC-D4PG) on "solvable constraint tasks"; and (2) obtains an improved trade-off between maximizing return and minimizing constraint overshoot on "unsolvable constraint tasks" compared to the baselines. (This includes a hard-constrained RL algorithm where the return simply collapses in such a case). \mglr achieves this by adapting the learning rate for the Lagrange multiplier update. This acts as a proxy for adapting the lagrangian multiplier. By amplifying/dampening the gradient updates to the lagrangian during training, the agent is able to influence the tradeoff between maximizing return and satisfying the constraints to yield the behavior of (1) and (2). We also implemented a meta-gradient approach called \mgrr that scales and offsets the shaped rewards. This approach did not outperform \mglr but is a direction of future work. The algorithm, derived meta-gradient update and a comparison to \mglr can be found in the Appendix, Section \ref{app:mgll_overview}. We show that across safety coefficients, domains and constraint thresholds, \mglr outperforms all of the baseline algorithms. We also derive the meta-gradient updates for \mglr and  perform an investigative study where we provide empirical intuition for the derived gradient update that helps explain this meta-gradient variant's performance. We believe the proposed techniques will generalize to other policy gradient algorithms but leave this for future work.

\bibliography{main}
\bibliographystyle{iclr2021_conference}
\newpage
\appendix
\newpage
\appendix

\section{Meta-Gradient for learning the Lagrange multiplier learning rate (\mglr)}\specified{Page 4 line 131}\label{sec:metal_derivation}%
This section details the meta-gradient derivation in \mglr for the meta-parameter $\eta=\lrlambda$.

\emph{Step 1}: Compute the inner loss updates for the Lagrange multiplier and the actor/critic networks.

Define the Lagrange multiplier loss as:

\begin{equation}
    L_{lambda} = \exp(\lrlambda) \lambda \biggl( \beta - \mathbb{E}\biggl[ C(s,a) | s=s_t, a = \pi_{\theta}(s_t) \biggr] \biggr)
\end{equation}

To update the Lagrange multiplier, compute $f(\tau,\lambda,\eta) = \alpha_1 \nabla_{\lambda} L_{lambda}(\lambda)$:

\begin{eqnarray*}
  f(\tau,\lambda,\eta) = \alpha_1 \nabla_{\lambda} L_{lambda}(\lambda) &=& \alpha_1 \exp(\lrlambda) \biggl( \beta - \mathbb{E}\biggl[ C(s,a) | s=s_t, a = \pi_{\theta}(s_t) \biggr] \biggr)
\end{eqnarray*}

The Lagrange multiplier is updated as follows:
\begin{equation}
    \lambda' = \lambda - f(\tau,\lambda, \eta)
    \label{eqn:lambda}
\end{equation}

Next, define the actor loss as:

\begin{equation}
    L_{actor}(\theta_{a}) = \Vert SG(\nabla_{a} Q_{\theta_{c}}(s_t,a_t) + a_{\theta_{a},t}) - a_{\theta_{a},t} \Vert^2_2
\end{equation}

where SG is a stop-gradient applied to the target.

To update the actor, compute $f(\tau,\theta_a,\eta) = -\alpha_{\theta_{a}} \nabla_{\theta_{a}} L_{actor}(\theta_a)$:

\begin{eqnarray}
  f(\tau,\theta_a, \eta) &=& \alpha_{\theta_{a}} \mathbb{E}\biggl[ \nabla_{a} Q(s,a)  \nabla_{\theta_a} \pi(s) \biggr]
\end{eqnarray}

\begin{equation}
    \theta_a' = \theta_a + f(\tau,\theta_a, \eta)  %
\end{equation}

The final inner update is the critic. Define the critic loss as the squared TD error:

\begin{equation}
    L_{critic} = (r(s,a) - \lambda' c(s,a) + \gamma Q_{T}(s',\pi_{T}(s')) - Q_{\theta_c}(s,a) )^2
\end{equation}

Note that the updated Lagrange multiplier $\lambda'$ is used in the critic loss computation. We therefore need to compute the gradient of the critic loss with respect to the critic parameters to be used in the critic update: $f(\tau,\theta_c, \lambda', \eta) = \alpha_{\theta_{c}} \nabla_{\theta_c} L_{critic}(\theta_c)$. 

Thus, the critic update is
\begin{equation*}
  f(\tau,\theta_c, \lambda', \eta) = \alpha_{\theta_{c}} \nabla_{\theta_c} L(\theta_c) =  -2 \alpha_{\theta_{c}} \cdot (r(s,a) - \lambda' c(s,a) + \gamma Q_{T}(s',\pi_{T}(s')) - Q_{\theta_c}(s,a)) \nabla_{\theta_{c}} Q_{\theta_c}(s,a)
\end{equation*}

\begin{equation}
    \theta_c' = \theta_c - f(\tau,\theta_c, \lambda', \eta)
    \label{eqn:critic_appendix}
\end{equation}

\emph{Step 2:} Define the meta-objective $J'(\tau', \theta'_c, \lambda'(\eta))$:

We define the meta-objective as $L_{outer}= L_{critic}$. That is,

\begin{equation}
    J'(\tau', \theta'_c (\lambda'(\eta)), \lambda'(\eta)) =  (r(s,a) - \lambda' c(s,a) + \gamma Q_{T}(s',\pi_{T}(s')) - Q_{\theta'_{c}}(s,a))^2
\end{equation}

\emph{Step 3:} Compute the meta-gradient:

$\frac{\partial J'(\tau',\theta'_{c}(\lambda'(\eta)),\lambda'(\eta))}{\partial \eta} =  (\nabla_{\theta'_{c}} J'(\tau',\theta'_{c}(\lambda'(\eta)),\lambda'(\eta)))^T \nabla_{\lambda'} \theta'_{c} \frac{\partial \lambda'}{\partial \eta} + \frac{\partial J'(\tau',\theta'_{c}(\lambda'(\eta)),\lambda'(\eta))}{\partial \lambda'}\frac{\partial \lambda'}{\partial \eta}$:

First start with $\frac{\partial \lambda'}{\partial \eta}$ and using Equation \ref{eqn:lambda}, we get:

\begin{eqnarray}
    \frac{d \lambda'}{d \eta} &=& \frac{d \lambda}{ d \eta} - \frac{\partial f(\tau,\lambda,\eta)}{\partial \eta}\\
    &=& - \frac{d f(\tau,\lambda,\eta)}{d \eta}= - \frac{\partial f(\tau,\lambda,\eta)}{\partial \lrlambda}\\
    &=& - \frac{\partial \alpha_1 \exp(\lrlambda) \biggl(\beta - \mathbb{E}\biggl[ C(s,a) | s=s_t, a =\pi_{\theta}(s_t) \biggr] \biggr)}{\partial \lrlambda} \\
    &=& \alpha_1 \exp(\lrlambda) \biggl(\mathbb{E}\biggl[ C(s,a) | s=s_t, a =\pi_{\theta}(s_t) \biggr] - \beta \biggr)\\
    &=& \alpha_1 \exp(\lrlambda) \biggl(J_C^{\pi} - \beta \biggr)
\end{eqnarray}

Then, compute $\frac{\partial J'(\tau',\theta'_{c}(\lambda'(\eta)),\lambda'(\eta))}{\partial \lambda'}$ which yields:

\begin{equation}
    \frac{\partial J'(\tau',\theta'_{c}(\lambda'(\eta)),\lambda'(\eta))}{\partial \lambda'} = -2 \cdot (r(s,a) - \lambda' c(s,a) + \gamma Q_{T}(s',\pi_{T}(s')) - Q_{\theta'_{c}}(s,a)) \cdot c(s,a)
\end{equation}

Compute $\nabla_{\theta'_{c}} J'(\tau',\theta'_{c}(\lambda'(\eta)),\lambda'(\eta))$:

\begin{equation}
    \nabla_{\theta'_{c}} J'(\tau',\theta'_{c}(\lambda'(\eta)),\lambda'(\eta)) = -2 \cdot (r(s,a) - \lambda' c(s,a) + \gamma Q_{T}(s',\pi_{T}(s')) - Q_{\theta'_{c}}(s,a))\nabla_{\theta'_{c}} Q_{\theta'_{c}}(s,a)
\end{equation}

Denote:

\begin{equation}
    \delta = r(s,a) - \lambda' c(s,a) + \gamma Q_{T}(s',\pi_{T}(s')) - Q_{\theta'_{c}}(s,a)
\end{equation}

Finally, compute $\nabla_{\lambda'} \theta'_{c}$. Using Equation \ref{eqn:critic_appendix}, we get:

\begin{eqnarray*}
   \nabla_{\lambda'} \theta'_{c} &=&  \nabla_{\lambda'} \theta_{c} -  \nabla_{\lambda'} f(\tau, \theta_c, \lambda', \eta)\\
   &=& - \nabla_{\lambda'} (-2 \alpha_{\theta_{c}} \cdot (r(s,a) - \lambda' c(s,a) + \gamma Q_{T}(s',\pi_{T}(s')) - Q_{\theta_c}(s,a)) \nabla_{\theta_{c}} Q_{\theta_c}(s,a) ) \\
   &=& - 2 \alpha_{\theta_{c}} \cdot c(s,a) \cdot  \nabla_{\theta_{c}} Q_{\theta_c}(s,a)
\end{eqnarray*}

The meta-gradient $\frac{\partial J'(\tau',\theta'_{c}(\lambda'(\eta)),\lambda'(\eta))}{\partial \eta}$ is therefore:

\begin{eqnarray*}
  &\frac{\partial J'(\tau',\theta'_{c}(\lambda'(\eta)),\lambda'(\eta))}{\partial \eta}&\\
  &=&  (\nabla_{\theta'_{c}} J'(\tau',\theta'_{c}(\lambda'(\eta)),\lambda'(\eta)))^T \nabla_{\lambda'} \theta'_{c} \frac{\partial \lambda'}{\partial \eta} + \frac{\partial J'(\tau',\theta'_{c}(\lambda'(\eta)),\lambda'(\eta))}{\partial \lambda'}\frac{\partial \lambda'}{\partial \eta}\\
  &=& -2 \delta \cdot (\nabla_{\theta'_{c}} Q_{\theta'_{c}}(s,a))^T \cdot  \biggl(-2 \alpha_{\theta_{c}}  \cdot c(s,a) \cdot  \nabla_{\theta_{c}} Q_{\theta_c}(s,a) \biggr) \cdot \biggl(\alpha_1 \exp(\lrlambda) \biggl(J_C^{\pi} - \beta \biggr) \biggr)\\
  &+& \biggl(-2 \delta \cdot c(s,a) \biggr) \cdot \biggl( \alpha_1 \exp(\lrlambda) \biggl(J_C^{\pi} - \beta \biggr) \biggr)\\
  &=& -2\delta \cdot c(s,a) \cdot \alpha_1 \exp(\lrlambda) \cdot\biggl(J_C^{\pi} - \beta \biggr) \biggl( -2 \alpha_{\theta_{c}}  (\nabla_{\theta'_{c}} Q_{\theta'_{c}}(s,a))^T \nabla_{\theta_{c}} Q_{\theta_c}(s,a) + 1 \biggr)\\
\end{eqnarray*}

The meta-gradient update is therefore:
\begin{eqnarray*}
  \lrlambda' &=& \lrlambda - \alpha_{\eta} \biggl( -2\delta \cdot c(s,a) \cdot \alpha_1 \exp(\lrlambda) \cdot\biggl(J_C^{\pi} - \beta \biggr) \biggl( -2 \alpha_{\theta_{c}}  (\nabla_{\theta'_{c}} Q_{\theta'_{c}}(s,a))^T \nabla_{\theta_{c}} Q_{\theta_c}(s,a) + 1 \biggr)\biggr)  \\
\end{eqnarray*}

\section{Meta-Gradient Reward Shaping (\mgrr)}
\label{app:mgll_overview}

We present \mgrr, an alternative approach to performing soft-constrained optimizing using meta-gradients. This approach predicts a scale and offset, $\kappa_{S}, \kappa_{O} = f_\phi(s_t, a_t, r_t, c_t)$ where $f_\phi$ is the meta-shaping network, with meta-parameters $\eta=\phi$, and the tuple $\langle s_t, a_t, r_t, c_t \rangle$ corresponds to the state, action, reward and per-step constraint-violation penalty $c_t$. These parameters shape the reward function to yield the meta-shaped reward: ${r_{meta-shaped}= r_t - \kappa_S \cdot \lambda c_t + \kappa_O}$. They provide additional degrees of freedom to shaping the reward which may yield an improved overall solution.

\mgrr trains two critics, an inner critic with parameters $\theta_{c,in}$ used in the inner loss and an outer critic with parameters $\theta_{c,out}$ which is used in the outer loss. The inner critic is trained using the meta-shaped reward $r_{meta-shaped}$ and the outer critic is trained using the regular shaped reward ${r_{shaped} = r_t - \lambda c_t}$. The Lagrange multiplier $\lambda$ is fixed for training the inner critic and is updated according to the Lagrange multiplier update equation for training the outer critic.

 The inner loss is defined as $L_{inner}(\theta_{a}, \theta_{c}, \lambda) = L_{actor}(\theta_{a}) + L_{critic, in}(\theta_{c}, \lambda)$ where $L_{actor}(\theta_{a}, \theta'_{c,in}) = -\mathbb{E}[Q_{\theta'_{c,in}}(s,\pi_{\theta_{a}}(s))]$ and $L_{critic, in}(\theta_{c}, \lambda) = \hat{\delta}^2$ where $\hat{\delta}$ is the squared TD error trained on the meta-shaped reward $r_{meta-shaped}$. The outer loss is $L_{outer}= J'(\theta'_{a},\theta'_{c,out}) =  \Vert SG(\nabla_{a} Q_{\theta'_{c,out}}(s_t,a_t) + a_{t,\theta'_{a}}) - a_{t, \theta'_{a}} \Vert^2$ which is the original D4PG actor loss. 

The full algorithmic overview is presented as Algorithm \ref{alg:mrrcd4pg}. We also derived the meta-gradient update and present it as Theorem \ref{thm:2} where we also provide intuition for the update itself. We also analyze the algorithm's performance under different hyper-parameter settings and meta-shaping reward formulations.

\subsection{Algorithm\specified{Page 5 line 152}}
The pseudo-code for \mgrr is shown in Algorithm~\ref{alg:mrrcd4pg}.

\begin{algorithm}[H]
\caption{\mgrr\label{alg:mrrc}}
\begin{algorithmic}[1]
\State \textbf{Input:} penalty $c(\cdot)$, constraint $C(\cdot)$, threshold $\beta$, learning rates $\alpha_1, \alpha_{\theta_{a}} ,\alpha_{\theta_{c,in}}, \alpha_{\theta_{c,out}}$, max.\ number of episodes $M$, fixed $\hat{\lambda}$ for inner-critic, meta-parameters $\eta=\phi$ (meta-shaping network parameters)
\State Initialize actor parameters $\theta_{a}$, inner critic parameters $\theta_{c,in}$, outer critic parameters $\theta_{c,out}$, Lagrange multiplier $\lambda = 0$
    \For{$1 \ldots M$}      
        
        \State Sample episode penalty $J_C^{\pi}$ from the penalty replay buffer
        \State $\lambda' \gets  [\lambda - \alpha_1 \left(J_C^{\pi} - \beta \right)]_{+} $ \Comment{\textbf{Lagrange multiplier update}}
        \State Sample batch with tuples $\langle s_t, a_t, r_t, c_t \rangle_{t=1}^{T}$ from ER and split into training/validation sets
                
        \State \textbf{Inner loss:} Accumulate and apply actor and critics updates over training batch $T_{train}$
        \State $\nabla \theta_a = 0, \nabla \theta_{c, in} = 0, \nabla \theta_{c, out} = 0$
        
        \For{$t = 1 \ldots T_{train}$}
            \State $\kappa_{S}, \kappa_{O} = f_\phi(s_t,a_t,r_t, c_t)$
            
            \State $\hat R_{t,in} = r_t - \kappa_{S}\cdot\hat{\lambda}c_t + \kappa_{O} + \gamma \hat Q_{in}(\lambda, s_{t+1}, a_{t+1}\sim \pi_T(s_{t+1}); \theta_{c, in})$
            
            \State $\hat R_{t,out} = r_t - \lambda' c_t + \gamma \hat Q_{out}(\lambda, s_{t+1}, a_{t+1}\sim \pi_T(s_{t+1}); \theta_{c,out})$
            
            \State $\nabla \theta_{c, in} \peq \alpha_{\theta_{c,in}} \partial (\hat R_{t,in} - \hat Q_{in}(\lambda, s_t, a_t; \theta_{c, in}))^2  / \partial \theta_{c, in}$ \Comment{\textbf{Inner critic update}}
            
            \State $\nabla \theta_{c, out} \peq \alpha_{\theta_{c, out}} \partial (\hat R_{t, out} - \hat Q_{out}(\lambda, s_t, a_t; \theta_{c, out}))^2  / \partial \theta_{c, out}$ \Comment{\textbf{Outer critic update}}           
            
            \State $\nabla \theta_a \peq \alpha_{\theta_{a}}  \mathbb{E} [\nabla_a Q_{in}(s_t,a_t) \nabla_{\theta_{a}} \pi_{\theta_{a}}(s_t)|_{a_t=\pi(s_t)}]$ \Comment{\textbf{Actor update}}          
        \EndFor
        
        \State $\theta'_{c, in} \gets \theta_{c, in} - \frac{1}{T_{train}}\sum \nabla \theta_{c, in}$
        \State $\theta'_{c, out} \gets \theta_{c, out} - \frac{1}{T_{train}}\sum \nabla \theta_{c, out}$
        \State $\theta'_{a} \gets \theta_{a} + \frac{1}{T_{train}}\sum \nabla \theta_a$
        
        \State \textbf{Outer loss: } Compute outer loss and meta-gradient update using validation set $T_{validate}$:
        \State $\phi' = \phi - \alpha_\eta \frac{J'(\theta'_{c,out}, \theta'_{a}(\phi))}{\partial \phi}$  \Comment{\textbf{Meta-parameters update}}
        \State $\lambda\gets\lambda',\theta_{a}\gets\theta'_{a}, \theta_{c, in}\gets\theta'_{c, in}, \theta_{c, out}\gets\theta'_{c, out}, \phi \gets \phi'$
    \EndFor
    
\State \textbf{return} policy parameters $\theta_{a}$
\end{algorithmic}
\label{alg:mrrcd4pg}
\end{algorithm}

\subsection{Meta-gradient derivation\specified{Page 2, footnote; Page 5 line 153; need to provide intuition of the update here as well}\label{app:mesh_derivation}}%
We first state the theorem for the meta-gradient update in \mgrr and then derive it.

\begin{theorem}
Let $\beta \geq 0$ be a pre-defined constraint violation threshold and $\eta=\phi$ be the parameters of the neural network $f_\phi$ used for reward meta-shaping, then the meta-gradient update is,

\begin{eqnarray*}
    \eta' &=& \eta - \alpha_\eta \cdot \biggl[2 \alpha_{\theta_{a}} \alpha_{\theta_{c,in}}  \mathbb{E}[\nabla_{a} Q_{\theta'_{c,out}}(s,a) \nabla_{\theta'_{a}} \pi(s) |_{a=\pi(s)}] \biggl(\nabla_{\theta'_{c,in}}\mathbb{E}[\nabla_a Q_{\theta'_{c,in}}(s,a) \nabla_{\theta_{a}} \pi_{\theta_{a}}(s)]_{a=\pi(s)} \biggr) \\
   &\cdot& \nabla_{\theta_{c,in}} Q_{\theta_{c,in}}(s,a) \biggl( \hat{\lambda} c(s,a) \cdot \nabla_{\phi} \kappa_{S} - \nabla_{\phi} \kappa_{T} \biggr) \biggr]
\end{eqnarray*}

where $\alpha_\eta$ is the meta-parameter learning rate.
\label{thm:2}
\end{theorem}

\emph{Step 1:} Define the inner losses and compute the inner loss updates.
First, we define the losses. 

Define the Lagrange multiplier loss as:

\begin{equation}
    L_{lambda} = \lambda \biggl( \beta - \mathbb{E}\biggl[ C(s,a) | s=s_t, a = \pi_{\theta}(s_t) \biggr] \biggr)
\end{equation}

To update the Lagrange multiplier, compute $f(\tau,\lambda,\eta) = \alpha_1 \nabla_{\lambda} L_{lambda}(\lambda)$:
 
\begin{eqnarray}
  f(\tau,\lambda,\eta) = \alpha_1  \nabla_{\lambda} L_{lambda}(\lambda) &=& \alpha_1 \biggl( \beta - \mathbb{E}\biggl[ C(s,a) | s=s_t, a = \pi_{\theta}(s_t) \biggr] \biggr)
\end{eqnarray}

\begin{equation}
    \lambda' = \lambda - f(\tau,\lambda, \eta)
\end{equation}

The inner update is performed using an inner critic with parameters $\theta_{c,in}$ and a fixed $\hat{\lambda}$ whose loss is defined as:

\begin{equation}
    L_{in,critic} = \hat{\delta}^2 \enspace ,
\end{equation}
where $$\hat{\delta} = r(s,a) - \kappa_{S}\cdot \hat{\lambda} c(s,a) + \kappa_{T} + \gamma Q_{T}(s',\pi_{T}(s')) - Q_{\theta_{c,in}}(s,a)$$ is the meta-shaped TD error.

$\kappa_{S}$ and $\kappa_{T}$ are the scale and the offset predicted by the meta-shaping network $f_\phi$, parameterized by $\phi$. The parameters of the meta-shaping network are the meta-parameters: $\eta=\phi$.

The inner critic's parameters are updated as follows,
\begin{equation}
    \theta_{c,in}' = \theta_{c,in} + f(\tau,\theta_{c,in}, \eta)
    \label{eqn:int_critic}
\end{equation}

where the update function $f(\tau,\theta_{c,in}, \eta)$ is:
\begin{eqnarray*}
  f(\tau,\theta_{c,in}, \eta) &=&  -\alpha_{\theta_{c,in}} \nabla_{\theta_{c,in}} L_{in,critic}(\theta_{c,in})\\ 
  &=&  2 \alpha_{\theta_{c,in}} \hat{\delta} \nabla_{\theta_{c,in}} Q_{\theta_{c,in}}(s,a)\\
\end{eqnarray*}

The updated critic parameters $\theta_{c,in}'$ are used in the actor loss:

\begin{equation}
    L_{actor}(\theta_{a}) = -\mathbb{E}[Q_{\theta'_{c,in}}(s,\pi_{\theta_{a}}(s))]
\end{equation}

where the expectation is with respect to the state distribution $d(s)$.

To update the actor, compute $f(\tau,\theta_a,\eta) =  - \alpha_{\theta_{a}} \nabla_{\theta_{a}} L_{actor}(\theta_a)$:

\begin{eqnarray}
  f(\tau,\theta_a, \eta) &=& \alpha_{\theta_{a}} \mathbb{E}[\nabla_a Q_{\theta'_{c,in}}(s,a) \nabla_{\theta_{a}} \pi_{\theta_{a}}(s)|_{a=\pi(s)}]
\end{eqnarray}

\begin{equation}
    \theta_a' = \theta_a + f(\tau,\theta_a, \eta)
    \label{eqn:actor_rew}
\end{equation}

\emph{Step 2:} Define the meta-objective outer loss $J'(\theta'_a)$. 

The outer critic with parameters $\theta_{c,out}$ is used in the outer loss and is trained on the shaped reward; it does not take the meta-parameter network predictions $\kappa_{S}$ and $\kappa_{T}$ into account. The outer critic loss is defined as:

\begin{equation}
    L_{critic} = \delta^2 \enspace ,
\end{equation}

where $$\delta = r(s,a) -  \lambda' c(s,a)  + \gamma Q_{T}(s',\pi_{T}(s')) - Q_{\theta_{c,out}}(s,a)$$ is the TD error.

The outer critic update is therefore:

\begin{equation}
    \theta_{c,out}' = \theta_{c,out} + f(\tau,\theta_{c,out})
    \label{eqn:critic}
\end{equation}

The meta-objective \textbf{outer} loss is defined as:
\begin{equation}
    J'(\theta'_{a}) =  \Vert SG(\nabla_{a} Q_{\theta'_{c,out}}(s_t,a_t) + a_{t,\theta'_{a}}) - a_{t, \theta'_{a}} \Vert^2
\end{equation}

\emph{Step 3:} Calculate the gradient of the outer loss with respect to the meta-parameters:
\begin{equation}
\nabla_\eta J'(\theta'_a) = \nabla_{\theta'_{a}} J'(\theta'_{a}) \cdot \nabla_{\theta'_{c,in}} \theta'_{a} \cdot \biggl(\frac{\partial \theta'_{c,in}}{\partial {\kappa_{S}}}\cdot \nabla_{\phi} \kappa_{S} +   \frac{\partial \theta'_{c,in}}{\partial \kappa_{T}}\cdot \nabla_{\phi} \kappa_{T}   \biggr)
\end{equation}

The sum occurs due to the multi-variate chain rule as $ \theta'_{c,in}(\kappa_{S}(\phi), \kappa_{T}(\phi))$.

Next, we calculate the gradients individually. The first one is:
\begin{equation}
    \nabla_{\theta'_{a}} J'(\theta'_{a}) = - \mathbb{E}[\nabla_{a} Q_{\theta'_{c,out}}(s,a) \nabla_{\theta'_{a}} \pi(s) |_{a=\pi(s)}]
\end{equation}

For the next gradient, $\nabla_{\theta'_{c,in}} \theta'_{a}$, we use Equation \ref{eqn:actor_rew} and have:

\begin{eqnarray}
   \nabla_{\theta'_{c,in}} \theta_a' &=& \nabla_{\theta'_{c,in}} \theta_a + \nabla_{\theta'_{c,in}} f(\tau,\theta_a, \eta)\\
   &=& \nabla_{\theta'_{c,in}} f(\tau,\theta_a, \eta)
\end{eqnarray}

and 
\begin{eqnarray}
\nabla_{\theta'_{c,in}} f(\tau,\theta_a, \eta) &=& \alpha_{\theta_{a}}(\nabla_{\theta'_{c,in}}\mathbb{E}[\nabla_a Q_{\theta'_{c,in}}(s,a) \nabla_{\theta_{a}} \pi_{\theta_{a}}(s)]_{a=\pi(s)})
\end{eqnarray}

Therefore:

\begin{equation}
    \nabla_{\theta'_{c,in}} \theta_a' = \alpha_{\theta_{a}}(\nabla_{\theta'_{c,in}}\mathbb{E}[\nabla_a Q_{\theta'_{c,in}}(s,a) \nabla_{\theta_{a}} \pi_{\theta_{a}}(s)]_{a=\pi(s)})
\end{equation}

For $\frac{\partial \theta'_{c,in}}{\partial \kappa_{S} }$ we use equation \ref{eqn:int_critic} and have:

\begin{eqnarray}
 \frac{\partial \theta_{c,in}'}{\partial \kappa_{S}} &=&  \frac{\partial \theta_{c,in}}{\partial \kappa_{S}} + \frac{\partial f(\tau,\theta_{c,in}, \eta)}{\partial \kappa_{S}}\\
\end{eqnarray}

Note that $f(\tau,\theta_{c,in}, \eta) = 2 \alpha_{\theta_{c,in}} \hat{\delta} \nabla_{\theta_{c,in}} Q_{\theta_{c,in}}(s,a)$. Therefore:

\begin{eqnarray}
   \frac{\partial \theta_{c,in}'}{\partial \kappa_{S}} &=&  \frac{\partial (2 \alpha_{\theta_{c,in}} \hat{\delta} \nabla_{\theta_{c,in}} Q_{\theta_{c,in}}(s,a))}{\partial {\kappa_{S}}}\\
   &=&  -2 \alpha_{\theta_{c,in}} \hat{\lambda} c(s,a)  \nabla_{\theta_{c,in}} Q_{\theta_{c,in}}(s,a)
\end{eqnarray}

Finally, for $\frac{\partial \theta'_{c,in}}{\partial \kappa_{T}}$, we get:

\begin{eqnarray}
   \frac{\partial \theta_{c,in}'}{\partial \kappa_{T}} &=&  \frac{\partial (2 \alpha_{\theta_{c,in}} \cdot \hat{\delta} \nabla_{\theta_{c,in}}Q_{\theta_{c,in}}(s,a))}{\partial \kappa_{T}}\\
   &=&  2 \alpha_{\theta_{c,in}}  \nabla_{\theta_{c,in}} Q_{\theta_{c,in}}(s,a)
\end{eqnarray}

Putting all this together, the meta-gradient is:

\begin{eqnarray*}
   \nabla_\eta J'(\theta'_a) &=& \nabla_{\theta'_{a}} J'(\theta'_{a}) \cdot \nabla_{\theta'_{c,in}} \theta'_{a} \cdot \biggl(\frac{\partial \theta'_{c,in}}{\partial \kappa_{S} }\cdot \nabla_{\phi} \kappa_{S} +   \frac{\partial \theta'_{c,in}}{\partial \kappa_{T}}\cdot \nabla_{\phi} \kappa_{T}   \biggr)\\
   &=& -\mathbb{E}[\nabla_{a} Q_{\theta'_{c,out}}(s,a) \nabla_{\theta'_{a}} \pi(s) |_{a=\pi(s)}] (\alpha_{\theta_{a}}\nabla_{\theta'_{c,in}}\mathbb{E}[\nabla_a Q_{\theta'_{c,in}}(s,a) \nabla_{\theta_{a}} \pi_{\theta_{a}}(s)|_{a=\pi(s)}]) \\
   &\cdot& \biggl[ -2 \alpha_{\theta_{c,in}} \hat{\lambda} c(s,a)  \nabla_{\theta_{c,in}} Q_{\theta_{c,in}}(s,a) \cdot \nabla_{\phi} \kappa_{S}  +2 \alpha_{\theta_{c,in}}  \nabla_{\theta_{c,in}} Q_{\theta_{c,in}}(s,a) \cdot \nabla_{\phi} \kappa_{T} \biggr]\\
   &=& 2 \alpha_{\theta_{a}} \alpha_{\theta_{c,in}} \mathbb{E}[\nabla_{a} Q_{\theta'_{c,out}}(s,a) \nabla_{\theta'_{a}} \pi(s) |_{a=\pi(s)}] \biggl(\nabla_{\theta'_{c,in}}\mathbb{E}[\nabla_a Q_{\theta'_{c,in}}(s,a) \nabla_{\theta_{a}} \pi_{\theta_{a}}(s)|_{a=\pi(s)}] \biggr) \\
   &\cdot& \nabla_{\theta_{c,in}} Q_{\theta_{c,in}}(s,a)) \biggl[ \hat{\lambda} c(s,a) \cdot \nabla_{\phi} \kappa_{S} - \nabla_{\phi} \kappa_{T} \biggr]\\
\end{eqnarray*}

So, the meta-gradient update for the meta-network parameters is:

\begin{eqnarray*}
    \eta' &=& \eta - \alpha_\eta \nabla_\eta J'(\theta'_a)\\
    &=& \eta - \alpha_\eta \cdot \biggl[2 \underbrace{\alpha_{\theta_{a}} \alpha_{\theta_{c,in}}}_{A}  \underbrace{\mathbb{E}[\nabla_{a} Q_{\theta'_{c,out}}(s,a) \nabla_{\theta'_{a}} \pi(s) |_{a=\pi(s)}]}_{B} \underbrace{\biggl(\nabla_{\theta'_{c,in}}\mathbb{E}[\nabla_a Q_{\theta'_{c,in}}(s,a) \nabla_{\theta_{a}} \pi_{\theta_{a}}(s)]_{a=\pi(s)} \biggr)}_{C} \\
   &\cdot& \underbrace{\nabla_{\theta_{c,in}} Q_{\theta_{c,in}}(s,a)}_{D} \underbrace{\biggl( \hat{\lambda} c(s,a) \cdot \nabla_{\phi} \kappa_{S} - \nabla_{\phi} \kappa_{T} \biggr)}_{E} \biggr]
\end{eqnarray*}

\subsubsection{Theoretical intuition}
We have divided the gradient into a number of terms (A, B, C, D, E) which we will now discuss to provide some intuition for the gradient update. The first term $A$ consists of the actor learning rate $\alpha_{\theta_{a}}$ and the inner critic learning rate $\alpha_{\theta_{c,in}}$. These terms significantly influence the magnitude of the meta-gradient update. We observed this effect in the meta-gradient experiments. 
The second term $B$ contains the outer loss deterministic policy gradient update, with respect to the updated parameters of the outer critic $\theta'_{c,out}$ and the actor $\theta'_{a}$ respectively.
Term $C$ corresponds to the inner loss deterministic policy gradient update with respect to the updated inner critic parameters $\theta'_{c,in}$ and the actor parameters $\theta_a$ from the previous update respectively. A gradient of this term is taken with respect to $\theta'_{c,in}$ which occurs since $\theta'_{c,in}$ is a function of $\phi$, the parameters of the meta-shaping network. 
Term $D$ corresponds to the gradient of the inner critic action value function with respect to the inner critic parameters. 
Term $E$ contains the value of the fixed Lagrangian multiplier $\hat{\lambda}$, the immediate cost $c(s,a)$ and the gradients of the scale and offset parameters, $\kappa_S$ and $\kappa_T$ with respect to the meta-network parameters $\phi$ respectively. If a constraint is violated, $c(s,a)=1$ then the Lagrangian multiplier, along with the scaling gradient, $\nabla_\phi \kappa_S$ has an effect on the overall meta-gradient update. This ensures that the meta-parameters are aware that the constraint has been violated. If the constraint is not violated, $c(s,a)=0$ ensuring that the Lagrangian multiplier does not need to be taken into account when updating the meta-network parameters. However, the offset parameter gradient $\nabla_\phi \kappa_T$ still has an effect on the overall meta-gradient update. This provides the agent with an additional degree of freedom for shaping the immediate reward.

\subsection{Implementation details and additional experiments}

As described in the main text, \mgrr shapes the reward using a scale ($\kappa_{S}$) and an offset ($\kappa_{O}$) predicted by a meta-shaping network $f_\phi$: ${r_\mathrm{meta-shaped} = r_t - \kappa_S \hat{\lambda} c_t + \kappa_O}$. The meta-shaping network, $f_\phi$, receives as input the per-timestep state, action, reward and penalty: $(s_t, a_t, r_t, c_t)$. $\hat{\lambda}$ is fixed at $0.1$.

The scale $\kappa_{S}$ is parameterized by a linear layer with a scaled sigmoid activation, $\kappa_{S} = \upsilon_S \cdot \sigma(\bar{\kappa}_S)$, where $\bar{\kappa}_S$ is the raw scale predicted by the meta-shaping network; hence, the scale $\kappa_{S}$ is bounded in the range $\left(0.0, \upsilon_S\right)$.

The offset $\kappa_{S}$ is parameterized by a linear layer with a scaled hyperbolic tangent activation, $\kappa_{O} = \upsilon_O \cdot \tanh(\bar{\kappa}_O)$, where $\bar{\kappa}_O$ is the raw offset predicted by the meta-shaping network; hence the offset $\kappa_{O}$ is bounded in the range $\left(-\upsilon_O, \upsilon_O\right)$.

Besides varying the architecture size (as mentioned in the main text), we also conducted experiments to analyze the effect on performance of an alternative formulation for the meta-shaped reward, $\kappa_S(r_t - \lambda c_t) + \kappa_O$. Performance of various \mgrr variants in terms of the reward formulation and the scaling factors for the raw predicted offset and scale are shown in Table~\ref{tab:mesh_extra}. The table shows that for the alternative reward shaping formulation, $\upsilon_S$, the multiplicative factor for $\kappa_S$, affects performance considerably, with a low value ($\upsilon_S=1.0$) obtaining considerably better reward and overshoot performance than a high value ($\upsilon_S=10.0$). In the main text we present the \mgrr variant which performed best, i.e.\ with the $r_\mathrm{meta-shaped}$ reward formulation, $\upsilon_O=3.0$, $\upsilon_S=10.0$, having an effective predicted Lagrange multiplier (i.e.\ $\kappa_{S}\cdot\hat{\lambda}$) in the range $\left(0.0, 1.0\right)$.

\begin{table}[!htb]
\begin{center}
\begin{tabular}{lllccc}
\toprule
Reward formulation & $\upsilon_O$ & $\upsilon_S$ &        $R$ &  $R_{penalized}$ &  $\max(0, J_C^\pi - \beta)$ \\
\midrule
\multirow{4}{*}{$\kappa_S(r_t - \lambda c_t) + \kappa_O$} & \multirow{2}{*}{1.0} & 1.0  &   738.02 $\pm$ 6.73 &           535.86 &                        0.20 \\
                                        &     & 10.0 &  140.62 $\pm$ 35.08 &          -145.48 &                        0.29 \\
\cmidrule{2-6}
                                        & \multirow{2}{*}{3.0} & 1.0  &   717.11 $\pm$ 8.00 &           542.35 &                        0.17 \\
                                        &     & 10.0 &  162.56 $\pm$ 19.65 &          -200.28 &                        0.36 \\
\cmidrule{1-6}
\multirow{2}{*}{$r_t - \kappa_S \lambda c_t + \kappa_O$} & 1.0 & 10.0 &   493.57 $\pm$ 0.94 &           338.85 &                        0.15 \\
                                        & 3.0 & 10.0 &   838.72 $\pm$ 6.36 &           438.74 &                        0.40 \\
\bottomrule
\end{tabular}
\end{center}
\caption{Average reward (and std.\ dev.\ over last 100 episodes at convergence), penalized reward and constraint overshoot for \mgrr variants on cartpole (single seed, 0.2 and 0.1 safety coefficients).\label{tab:mesh_extra}}
\end{table}

As shown in Figure 5, \mglr outperforms different variants of \mgrr. 

\begin{figure}[!htb]%
\centering
        \centering
        \includegraphics[width=\linewidth]{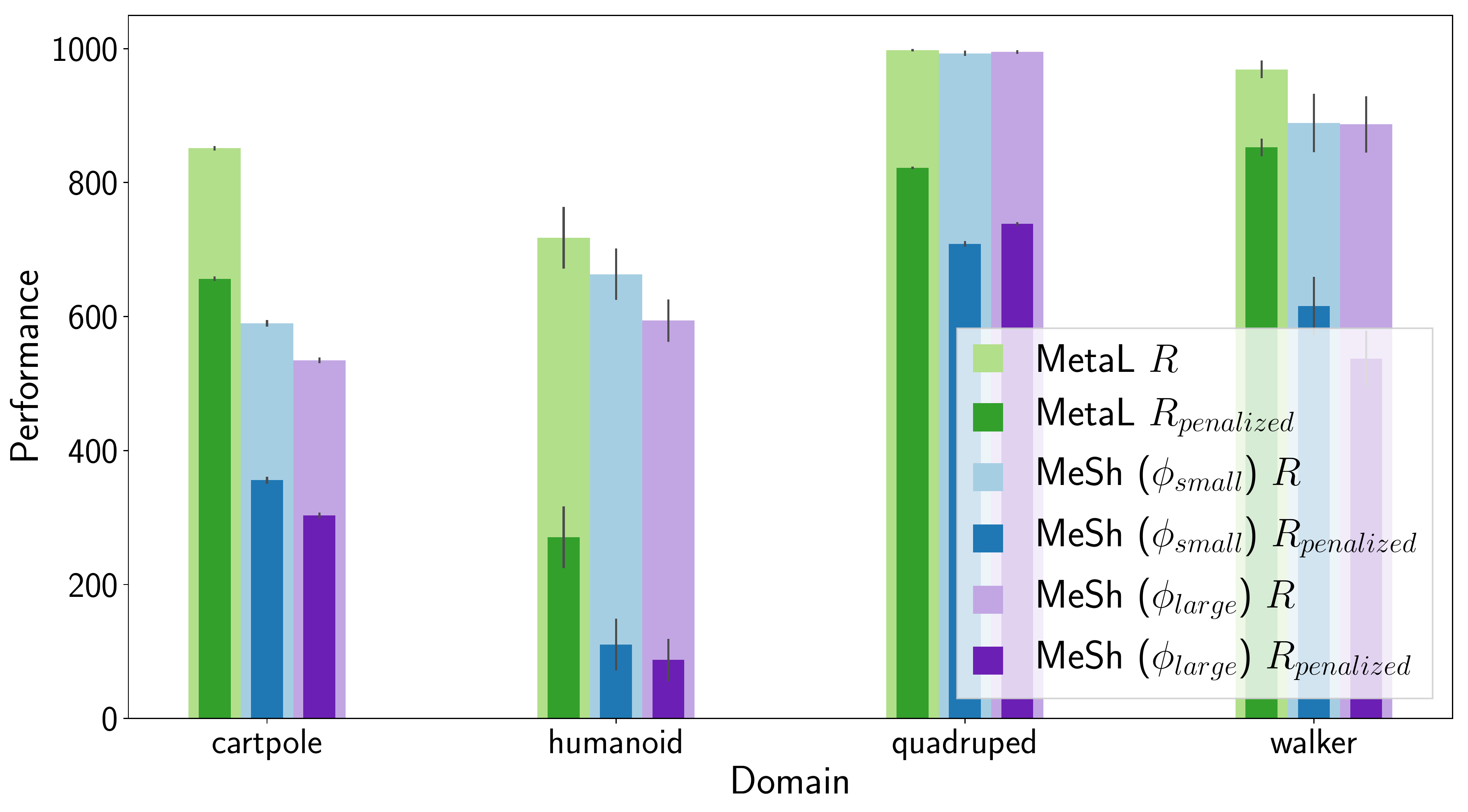}
  \caption{\mglr and \mgrr performance comparison across domains. As seen in the figure, \mglr clearly outperforms \mgrr. }
  \label{app:metalmesh_fig}
  \vspace{-0.2cm}
\end{figure}

\section{Constraint Definitions}\label{app:constraint_defs}\specified{Page 5 line 162}%
The per-domain safety constraints of the Real-World Reinforcement Learning (RWRL) suite that we use in the paper are given in Table~\ref{tab:safe_envs_full}.

\begin{table}
\small
\begin{center}
\begin{tabular}{p{4.5cm} p{3.5cm}}
 \multicolumn{2}{l}{\textbf{Cartpole} variables: $x, \theta$} \\
 \hline
 {Type} & {Constraint} \\
 \hline
 \texttt{slider\_pos} & 
  $x_l < x < x_r$ \\
 \texttt{slider\_accel}
 &$ \ddot{x} < A_\textit{max}$ \\
 \texttt{balance\_velocity}*
 & $ \left| \theta \right| > \theta_L \vee \dot{\theta} < \dot{\theta}_V $ \\
 \hline
\end{tabular}
\\
\begin{tabular}[t]{p{4.5cm} p{3.5cm}}
\multicolumn{2}{l}{} \\
\multicolumn{2}{l}{\textbf{Walker} variables: $\bm{\theta}, \bm{u}, \bm{F}$} \\
\hline
 {Type} & {Constraint} \\
 \hline
 \texttt{joint\_angle}
 & $\bm{\theta}_{L} < \bm{\theta} < \bm{\theta}_{U}$ \\
 \texttt{joint\_velocity}*
 & $ \max_i \left| \dot{\bm{\theta}_i} \right| < L_{\dot{\theta}} $ \\
 \texttt{dangerous\_fall}
 & $0 < (\bm{u}_z \cdot \bm{x})$ \\
 \texttt{torso\_upright}
  &  $0 < \bm{u}_z$\\
 \hline
\end{tabular}
\\
\begin{tabular}[t]{p{4.5cm} p{3.5cm}}
\multicolumn{2}{l}{} \\
\multicolumn{2}{l}{\textbf{Quadruped} variables: $\bm{\theta}, \bm{u}, \bm{F}$} \\
\hline
 {Type} & {Constraint} \\
 \hline
  \texttt{joint\_angle}*
 & $\theta_{L, i} < \bm{\theta}_i < \theta_{U,i}$ \\
 \texttt{joint\_velocity}
 & $ \max_i \left| \dot{\bm{\theta}_i} \right| < L_{\dot{\theta}} $ \\
 \texttt{upright}
 &  $0 < \bm{u}_z$\\
\texttt{foot\_force}
 &$ \bm{F}_{\textit{EE}} < F_\textit{max} $ \\
\hline
\end{tabular}
\\
\begin{tabular}[t]{p{4.5cm} p{3.5cm}}
 \multicolumn{2}{l}{} \\
\multicolumn{2}{l}{\textbf{Humanoid} variables: $\bm{\theta}, \bm{u}, \bm{F}$} \\
\hline
 {Type} & {Constraint} \\
 \hline
 \texttt{joint\_angle\_constraint}*
 & $\theta_{L, i} < \bm{\theta}_i < \theta_{U,i}$ \\
 \texttt{joint\_velocity\_constraint}
 & $\max_i \left| \dot{\bm{\theta}_i} \right| < L_{\dot{\theta}}$ \\
 \texttt{upright\_constraint} 
 & $0 < \bm{u}_z$ \\
 \multirow{2}{*}{\texttt{dangerous\_fall\_constraint}}
 & $ \bm{F}_{head} < F_{\textit{max},1}$ \\
 & $ \bm{F}_{torso} < F_{\textit{max},2}$ \\
 \texttt{foot\_force\_constraint} 
 & $ \bm{F}_{\textit{Foot}} < F_\textit{max,3} $ \\
 \hline

\end{tabular}
\end{center}
\caption{Safety constraints available for each RWRL suite domain; the constraints we use in this paper are indicated by an asterisk (*).}
\label{tab:safe_envs_full}
\end{table}

\section{Experimental Setup}
The action and observation dimensions for the task can be seen in Table \ref{tab:state_dimensions}. Parameters that were used for each variant of D4PG can be found in Table~\ref{tab:d4pg_hyperparameters}. The per-domain constraint thresholds ($\beta$) can be found in Table~\ref{tab:thresholds}.

\begin{table}[htpb]
\begin{center}
\begin{tabular}{l c c}
\toprule
{Domain: Task}             & {Observation Dimension} & {Action Dimension} \\
\midrule
{Cartpole: Swingup} & 5                        & 1                         \\ %
{Walker: Walk}      & 18                       & 6                         \\ %
{Quadruped: Walk}   & 78                       & 12                          \\ %
{Humanoid: Walk}    & 67                       & 21                        \\
\bottomrule
\end{tabular}
\end{center}
\caption{Observation and action dimension for each task.\label{tab:state_dimensions}}
\end{table}

\begin{table}[htpb]
\begin{center}
 \begin{tabular}{l c} 
 \toprule
 Hyperparameter & Value \\
 \midrule
 Policy net & 256-256-256 \\ 
 $\sigma$ (exploration noise) & 0.1 \\
 Critic net & 512-512-256 \\ 
 Critic num. atoms & 51\\
 Critic vmin & -150 \\
 Critic vmax & 150\\ 
 N-step transition & 5\\
 Discount factor ($\gamma$) & 0.99 \\
 Policy and critic opt. learning rate & 0.0001 \\
 Replay buffer size & 1000000 \\
 Target network update period & 100\\
 Batch size & 256\\
 Activation function & elu\\
 Layer norm on first layer & Yes\\
 Tanh on output of layer norm & Yes\\
 \bottomrule
\end{tabular}
\end{center}
\caption{Hyperparameters for all variants of D4PG.\label{tab:d4pg_hyperparameters}}
\end{table}

\begin{table}[htpb]
\begin{center}
\begin{tabular}{l c}
\toprule
Domain    & Thresholds                        \\
\midrule
Cartpole  & 0.07, 0.09, 0.115 \\ %
Quadruped & 0.545, 0.645, 0.745 \\ %
Walker & 0.057, 0.077, 0.097 \\ %
Humanoid  & 0.278, 0.378, 0.478 \\
\midrule
\end{tabular}
\end{center}
\caption{Constraint thresholds ($\beta$) for each domain.\label{tab:thresholds}}
\end{table}

\section{Additional \mglr Experiments}

\subsection{Performance as a function of domain}
In Table~\ref{tab:domains} we present quantitative performance for all algorithms as a function of the domain.

\begin{table}
\centering
\begin{tabular}{llccc}
\toprule %
Domain &    Algorithm &  $R_{penalized}$ &    $R$ &  \footnotesize{$\max(0, J_C^\pi - \beta)$} \\
\midrule
Cartpole & D4PG &    458.58 $\pm$ 10.56\significant &  \textbf{859.18} &      0.40 \\
       & RC-D4PG &    504.46 $\pm$ 88.76\significant &           582.74 &      0.08 \\
       & \textbf{MetaL} &        \textbf{656.86 $\pm$ 6.30} &           853.13 &      0.20 \\
       & RS-0.1 &     648.64 $\pm$ 5.30\significant &           854.88 &      0.21 \\
       & RS-1.0 &     163.22 $\pm$ 0.37\significant &           163.22 &      0.00 \\
       & RS-10.0 &     162.35 $\pm$ 0.40\significant &           162.35 &      0.00 \\
       & RS-100.0 &     161.71 $\pm$ 0.42\significant &           161.71 &      0.00 \\\midrule
Humanoid & D4PG &    248.43 $\pm$ 10.25\significant &  \textbf{870.26} &      0.62 \\
       & RC-D4PG &  -170.22 $\pm$ 131.09\significant &           295.01 &      0.47 \\
       & \textbf{MetaL} &       \textbf{383.27 $\pm$ 54.57} &           857.06 &      0.47 \\
       & RS-0.1 &    258.09 $\pm$ 70.24\significant &           792.26 &      0.53 \\
       & RS-1.0 &    195.52 $\pm$ 11.50\significant &           662.02 &      0.47 \\
       & RS-10.0 &   -567.74 $\pm$ 78.24\significant &             1.38 &      0.57 \\
       & RS-100.0 &    -620.63 $\pm$ 0.35\significant &             1.30 &      0.62 \\\midrule
Quadruped & D4PG &     647.71 $\pm$ 0.22\significant &           999.06 &      0.35 \\       
       & RC-D4PG &               773.39 $\pm$ 109.53 &           905.62 &      0.13 \\
       & MetaL &                828.21 $\pm$ 19.73 &           998.89 &      0.17 \\
       & RS-0.1 &                818.45 $\pm$ 13.31 &  \textbf{999.40} &      0.18 \\
       & RS-10.0 &   631.53 $\pm$ 121.03\significant &           769.81 &      0.14 \\
       & RS-100.0 &   524.10 $\pm$ 226.46\significant &           687.90 &      0.16 \\
       & \textbf{RS-1.0} &       \textbf{836.96 $\pm$ 19.57} &           998.98 &      0.16 \\\midrule
Walker & D4PG &    376.09 $\pm$ 26.94\significant &  \textbf{982.14} &      0.61 \\       
       & RC-D4PG &    806.77 $\pm$ 27.67\significant &           810.32 &      0.00 \\
       & MetaL &                843.38 $\pm$ 22.51 &           975.56 &      0.13 \\
       & RS-0.1 &                840.45 $\pm$ 17.82 &           980.48 &      0.14 \\
       & RS-10.0 &    608.13 $\pm$ 46.19\significant &           608.13 &      0.00 \\
       & RS-100.0 &    408.80 $\pm$ 22.93\significant &           408.80 &      0.00 \\
       & \textbf{RS-1.0} &       \textbf{851.10 $\pm$ 30.55} &           912.97 &      0.06 \\
\bottomrule
\end{tabular}
\caption{Performance per domain. \mglr compared to baselines in terms of return, penalized return and constraint overshoot average across the safety coefficients and thresholds for each domain. \mglr is statistically significantly better than RS-0.1 (the best overall reward shaping baseline) in three out of four domains and is better than RS-1.0 in two out of four domains. Overall, \mglr is statistically significantly better than all the baselines across all domains, safety coefficients and thresholds with all p-values less than $1e-09$. }
\label{tab:domains}
\end{table}

\subsection{Sub-optimal Learning Rate Performance}\label{app:suboptimal}%
\begin{figure}[!htb]
    \centering
    \includegraphics[width=\linewidth]{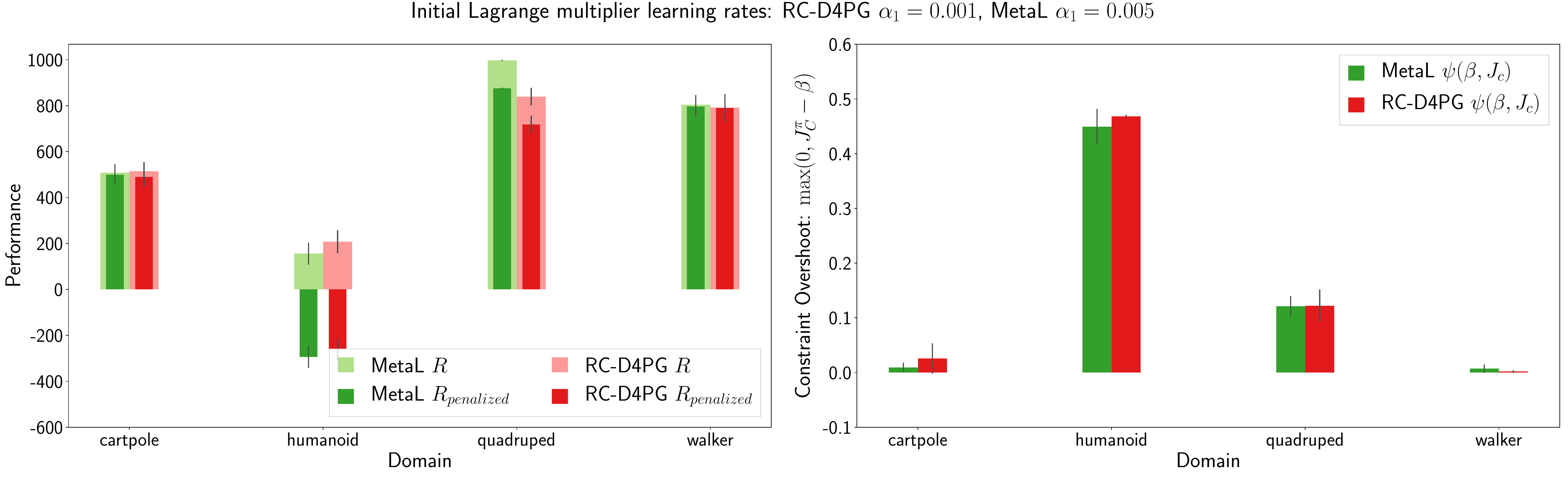}    
    \caption{Performance (left) and overshoot (right) as a function of the best performing Lagrange learning rate for \rcpo ($\alpha_1=0.001$) and the \emph{worst} performing initial Lagrange learning rate for \mglr ($\alpha_1=0.005$).}
    \label{fig:best_worst}    
\end{figure}

Another indication of the improved performance of \mglr comes from comparing its performance on the largest initial Lagrange learning rate (corresponding to its \emph{lowest} average performance) with \rcpo on the smallest learning rate (corresponding to its \emph{highest} average performance). The results can be found in Figure~\ref{fig:best_worst} for return and penalized return (left) and overshoot (right) respectively. We noticed two interesting results. First, the performance is similar (if not slightly better for \mglr) on a sub-optimal learning rate for both return and penalized return. Second, the overshoot performance is better for \mglr indicating that it learns to satisfy constraints better than \rcpo, even at this sub-optimal learning rate.

\label{app:experiments}

\subsection{Initial learning rate performance}\label{app:initiallearningrate}

\begin{figure}[!htb]
    \centering
    \includegraphics[width=\linewidth]{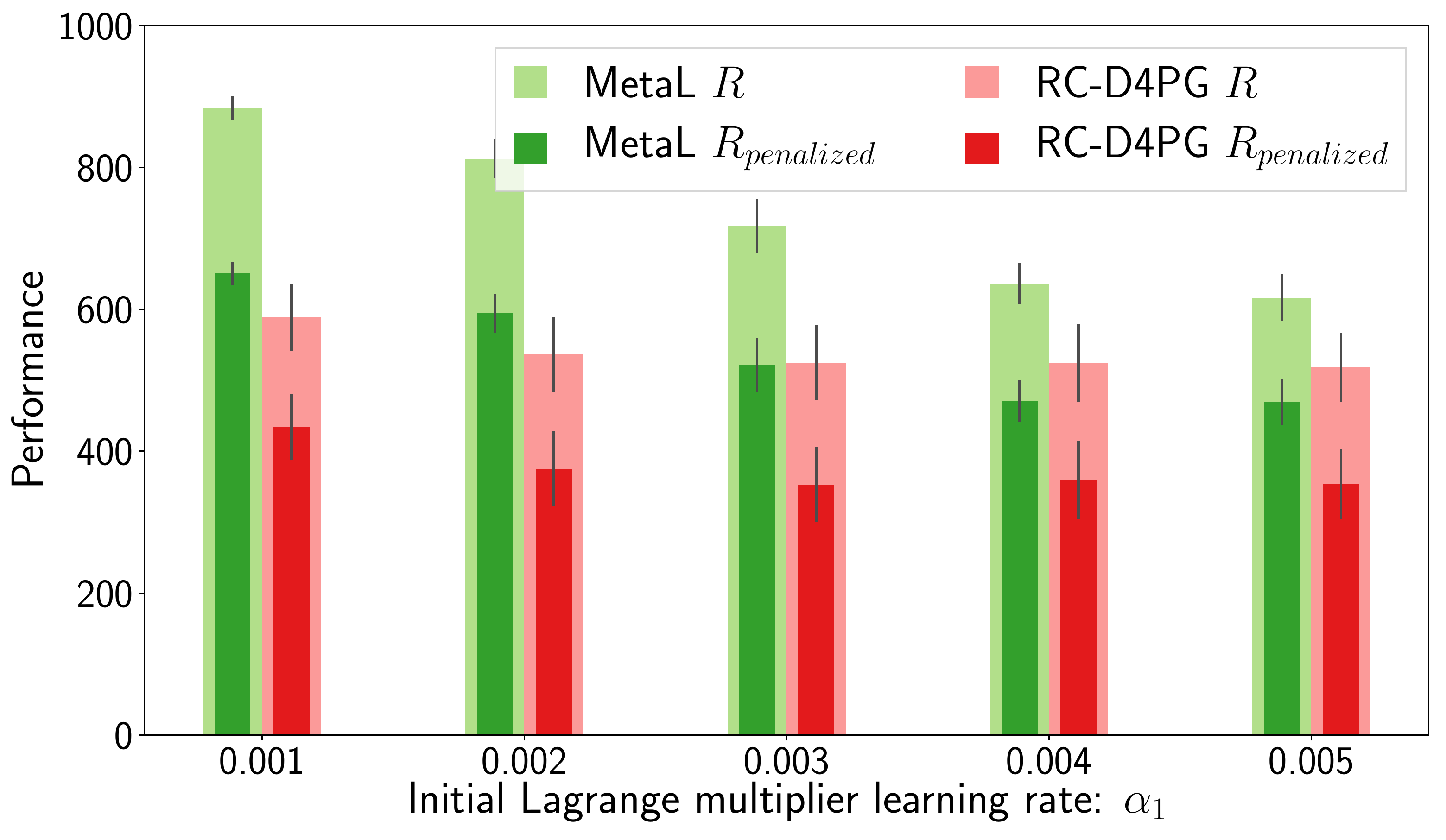}    
    \caption{The performance of \mglr compared to RC-D4PG in terms of initial fixed learning rates. Even though both \mglr and RC-D4PG are sensitive to the initial fixed learning rate, \mglr consistently outperforms RC-D4PG across all of the tested learning rates.}
    \label{fig:initial_learning_rate}    
\end{figure}

Figure \ref{fig:initial_learning_rate} plots the performance of \mglr and RC-D4PG as a function of the initial fixed Lagrange multiplier learning rate. Note that for RC-D4PG this learning rate is fixed throughout training. As seen in the figure, both methods are sensitive to the initial learning rate. However, \mglr consistently outperforms RC-D4PG for all of the tested fixed learning rates.

\section{Reward Constrained D4PG Algorithm}
The pseudo-code for \rcpo can be found in Algorithm~\ref{alg:rcpo}.

\begin{algorithm}[H]
\caption{Reward Constrained D4PG (RC-D4PG)}\label{alg:rcpo}
\begin{algorithmic}[1]
\State \textbf{Input:} penalty $c(\cdot)$, constraint $C(\cdot)$, threshold $\beta$, learning rates $\alpha_1 < \alpha_{\theta_a} < \alpha_{\theta_c}$, max.\ number of episodes $M$
\State Initialize actor parameters $\theta_a$, critic parameters $\theta_c$, Lagrange multipliers $\lambda = 0$
    \For{$1 \ldots M$}

        \State Sample episode penalty $J_C^{\pi}$ from the penalty replay buffer
        \State $\lambda \gets  [\lambda - \alpha_1 \left(J_C^{\pi} - \beta \right)]_{+} $ \Comment{\textbf{Lagrange multiplier update}}
        
        \State Sample batch $\langle s_t, a_t, r_t, s_{t+1}, c_t\rangle_{t=1}^{T}$ from ER
        \State $\nabla \theta_c = 0, \nabla \theta_a = 0$
        \For{$t = 1 \ldots T$}
            \State $\hat R_t = r_t - \lambda c_t + \gamma \hat Q(\lambda, s_{t+1}, a_{t+1}\sim \pi_{Target}(s_{t+1}); \theta_{c})$
            \State $\nabla \theta_c \peq \alpha_{\theta_{c}} \partial (\hat{R}_t - \hat Q(\lambda, s_t, a_t; \theta_{c}))^2 / \partial \theta_{c}$ \Comment{\textbf{Critic update}}
            \State $\nabla \theta_a \peq \alpha_{\theta_{a}}  \mathbb{E} [\nabla_a Q(s_t,a_t) \nabla_{\theta_{a}} \pi_{\theta_{a}}(s_t)|_{a_t=\pi(s_t)}]$ \Comment{\textbf{Actor update}}
            
        \EndFor
        \State $\theta_{c} \gets \theta_{c} - \frac{1}{T}\sum \nabla \theta_c$
        \State $\theta_{a} \gets \theta_{a} + \frac{1}{T}\sum \nabla \theta_a$        
    \EndFor
    
\State \textbf{return} policy parameters $\theta_a$
\end{algorithmic}
\label{alg:rcd4pg}
\end{algorithm}

\end{document}